\documentclass[sigconf]{acmart}

\usepackage{multirow}
\usepackage{makecell}
\usepackage{algpseudocode}
\usepackage{algorithm}
\usepackage{cleveref}
\usepackage{mathtools, amsthm}
\usepackage{blindtext}
\usepackage{microtype}
\usepackage{bm}
\usepackage{thmtools}
\usepackage{enumitem}
\newcommand{\lrbrac}[1]{\left(#1\right)}
\newcommand{\var}{\operatorname{Var}}
\newcommand{\cov}{\operatorname{Cov}}
\newcommand{\lrang}[1]{\left\langle#1\right\rangle}

\declaretheorem[
  shaded={rulecolor=black, rulewidth=0.0pt},
  name=Proposition,
]{proposition}

\AtBeginDocument{%
  }

\setcopyright{acmlicensed}
\copyrightyear{2018}
\acmYear{2018}
\acmDOI{XXXXXXX.XXXXXXX}

\acmConference[Conference acronym 'XX]{Make sure to enter the correct
  conference title from your rights confirmation emai}{June 03--05,
  2018}{Woodstock, NY}
\acmISBN{978-1-4503-XXXX-X/18/06}

\newcommand{\method}{\textsc{GROOT }}
\newcommand{\methodname}{\textsc{GROOT}}

\newcommand\norm[1]{\| #1 \|}

\newcommand\conv{\text{Conv}(\mathbb{X})}

\newcommand{\surrogate}{f_\Phi}

\begin{document}

\title{\methodname: Effective Design of Biological Sequences \\ with Limited Experimental Data}

\author{Thanh V. T. Tran $^\dagger$}
\affiliation{%
  \institution{FPT Software AI Center}
  \city{Hanoi}
  \country{Vietnam}
}
\email{ThanhTVT1@fpt.com}

\author{Nhat Khang Ngo $^\dagger$}
\affiliation{%
  \institution{FPT Software AI Center}
  \city{Ho Chi Minh}
  \country{Vietnam}}
\email{KhangNN4@fpt.com}

\author{Viet Anh Nguyen}
\affiliation{%
  \institution{FPT Software AI Center}
  \city{Hanoi}
  \country{Vietnam}
}
\email{AnhNV117@fpt.com}

\author{Truong Son Hy$^*$} \thanks{$\dagger$: Equal contribution. $*$: Corresponding Author.}
\affiliation{%
  \institution{The University of Alabama at Birmingham}
  \city{Birmingham}
  \state{AL 35294}
  \country{United States}
}
\email{thy@uab.edu}


\renewcommand{\shortauthors}{Trovato et al.}


\begin{abstract}
Latent space optimization (LSO) is a powerful method for designing discrete, high-dimensional biological sequences that maximize expensive black-box functions, such as wet lab experiments. This is accomplished by learning a latent space from available data and using a surrogate model $\surrogate$ to guide optimization algorithms toward optimal outputs. However, existing methods struggle when labeled data is limited, as training $\surrogate$ with few labeled data points can lead to subpar outputs, offering no advantage over the training data itself. We address this challenge by introducing \textbf{\method}, a \textbf{GR}aph-based Latent Sm\textbf{O}\textbf{O}\textbf{T}hing for Biological Sequence Optimization. In particular, \method generates pseudo-labels for neighbors sampled around the training latent embeddings. These pseudo-labels are then refined and smoothed by Label Propagation. Additionally, we theoretically and empirically justify our approach, demonstrate \methodname's ability to extrapolate to regions beyond the training set while maintaining reliability within an upper bound of their expected distances from the training regions. We evaluate \method on various biological sequence design tasks, including protein optimization (GFP and AAV) and three tasks with exact oracles from Design-Bench. The results demonstrate that \method equalizes and surpasses existing methods without requiring access to black-box oracles or vast amounts of labeled data, highlighting its practicality and effectiveness. We release our code at \url{https://anonymous.4open.science/r/GROOT-D554}.
\end{abstract} 




\maketitle

\section{Introduction} \label{sec:intro}

Proteins are crucial biomolecules that play diverse and essential roles in every living organism. Enhancing protein functions or cellular fitness is vital for industrial, research, and therapeutic applications \cite{Huang2016, doi:10.1021/acs.chemrev.1c00260}. One powerful approach to achieve this is \textit{directed evolution}, which involves iteratively performing random mutations and screening for proteins with desired phenotypes \cite{ARNOLD19965091}. However, the vast space of possible proteins makes exhaustive searches impractical in nature, the laboratory, or computationally. Consequently, recent machine learning (ML) methods have been developed to improve the sample efficiency of this evolutionary search \cite{pmlr-v162-jain22a, pmlr-v162-ren22a, lee2024robust}. When experimental data is available, these approaches use a surrogate model $\surrogate$, trained to guide optimization algorithms toward optimal outputs.

While these methods have achieved state-of-the-art results on various benchmarks, most neglect the scenario of extremely limited labeled data and fail to utilize the abundant unlabeled data. This is a significant issue because exploring protein function and characteristics through iterative mutation in wet-lab is costly \cite{Fowler2014}. As later shown in \Cref{sec:numerical-results}, previous works perform poorly in this context, proving that training $\surrogate$ with few labeled data points can lead to subpar designs, offering no advantage over the training data itself. One potential solution for dealing with noisy and limited data is to \textit{regularize} the fitness landscape model, smoothing the sequence representation and facilitating the use of gradient-based optimization algorithms. Although some works have addressed this problem (Section \ref{sec:related}), none can optimize effectively in the case of labeled scarcity.

\subsection{Contribution}
Our work builds on the observation by \citet{pmlr-v162-trabucco22a} that surrogate models $\surrogate$ trained on limited labeled data are vulnerable to noisy labels. This sensitivity can lead the model to sample false negatives or become trapped in suboptimal solutions (local minima) when used for guiding offline optimization algorithms. To address this issue, the ultimate objective is to enhance the predictive ability of $\surrogate$, reducing the risk of finding suboptimal solutions. To fulfill this, we propose \textbf{\method}, a novel framework designed to tackle the problem of labeled scarcity by generating synthetic sample representations from existing data. From the features encoded by the encoder, we formulate sequences as a graph with fitness values as node attributes and apply Label Propagation \cite{NIPS2003_87682805} to generate labels for newly created nodes synthesized through interpolation within existing nodes. The smoothed data is then fitted with a neural network for optimization. 
Figure \ref{fig:method_overview} provides an overview of our framework. We evaluate our method on two fitness optimization tasks: Green Fluorescent Proteins (GFP) \cite{sarkisyan2016local} and Adeno-Associated Virus (AAV) \cite{bryant2021deep}. Our results show that \method outperforms previous state-of-the-art baselines across all difficulties of both tasks. We also demonstrate that our method performs stably in extreme cases where there are fewer than 100 labeled training sequences, unlike other approaches. Our contributions are summarized as follows:
\begin{itemize}
\item We introduce \method, a novel framework that uses graph-based smoothing to train a smoothed fitness model, which is then used in the optimization process.
\item We theoretically show that our smoothing technique can expand into extrapolation regions while keeping a reasonable distance from the training data. This helps to reduce errors occurring when the surrogate model makes predictions for unseen points too far from the training set. 


\item We empirically show that \method achieves state-of-the-art results across all difficulties in two protein design benchmarks, AAV and GFP. Notably, in extreme cases with very limited labeled data, our approach performs stably, achieving $6$ fold fitness improvement in GFP and $1.3$ times higher in AAV compared to the training set.

\item We further evaluate \method on diverse tasks of different domains (e.g., robotics, DNA) within the Design-Bench. Our experimental results are competitive with state-of-the-art approaches, highlighting the method's domain-agnostic capabilities.


\end{itemize}

\section{Related work} \label{sec:related}
\paragraph{\textbf{LSO for Sequence Design}} Directed evolution is a traditional paradigm in protein sequence design that has achieved notable successes \cite{https://doi.org/10.1002/anie.201708408}. Within this framework, various machine learning algorithms have been proposed to improve the sample efficiency of evolutionary searches \cite{pmlr-v162-ren22a, Qiu_2022, Emami2023, Tran2023.11.28.568945}. However, most of these methods optimize protein sequences directly in the sequence space, dealing with discrete, high-dimensional decision variables. Alternatively, \citet{doi:10.1021/acscentsci.7b00572} and \citet{Castro2022} employed a VAE model and applied gradient ascent to optimize the latent representation, which is then decoded into biological sequences. Similarly, \citet{lee2023protein} used an off-policy reinforcement learning method to facilitate updates in the representation space. Another notable approach \cite{pmlr-v162-stanton22a} involves training a denoising autoencoder with a discriminative multi-task Gaussian process head, enabling gradient-based optimization of multi-objective acquisition functions in the latent space of the autoencoder. Recently, latent diffusion has been introduced for designing novel proteins, leveraging the capabilities of a protein language model \cite{chen2023ampdiffusion}.

\paragraph{\textbf{Protein Fitness Regularization}} The NK model was an early effort to represent protein fitness smoothness using a statistical model of epistasis \cite{Kauffman1989}. \citet{doi:10.1021/acscentsci.7b00572} approached this by mapping sequences to a regularized latent fitness landscape, incorporating an auxiliary network to predict fitness from the latent space. \citet{Castro2022} further enhanced this work by introducing negative sampling and interpolation-based regularization into the regularization of latent representation. Additionally, \citet{frey2024protein} proposed to learn a smoothed energy function, allowing sequences to be sampled from the smoothed data manifold using Markov Chain Monte Carlo (MCMC). Currently, GGS \cite{kirjner2024improving} is most related to our work by utilizing discrete regularization using graph-based smoothing techniques. However, GGS enforces similar sequences to have similar fitness, potentially leading to suboptimal performance since fitness can change significantly with a single mutation \cite{doi:10.1073/pnas.2109649118}. Our work addresses this by encoding sequences into the latent space and constructing a graph based on the similarity of latent vectors. Furthermore, since our framework operates in the latent space, it is not constrained by biological domains and can be applied to other fields like robotics, as demonstrated in \Cref{app:design-bench}.

\section{Method}
\begin{figure*}
    \centering
    \includegraphics[width=\textwidth]{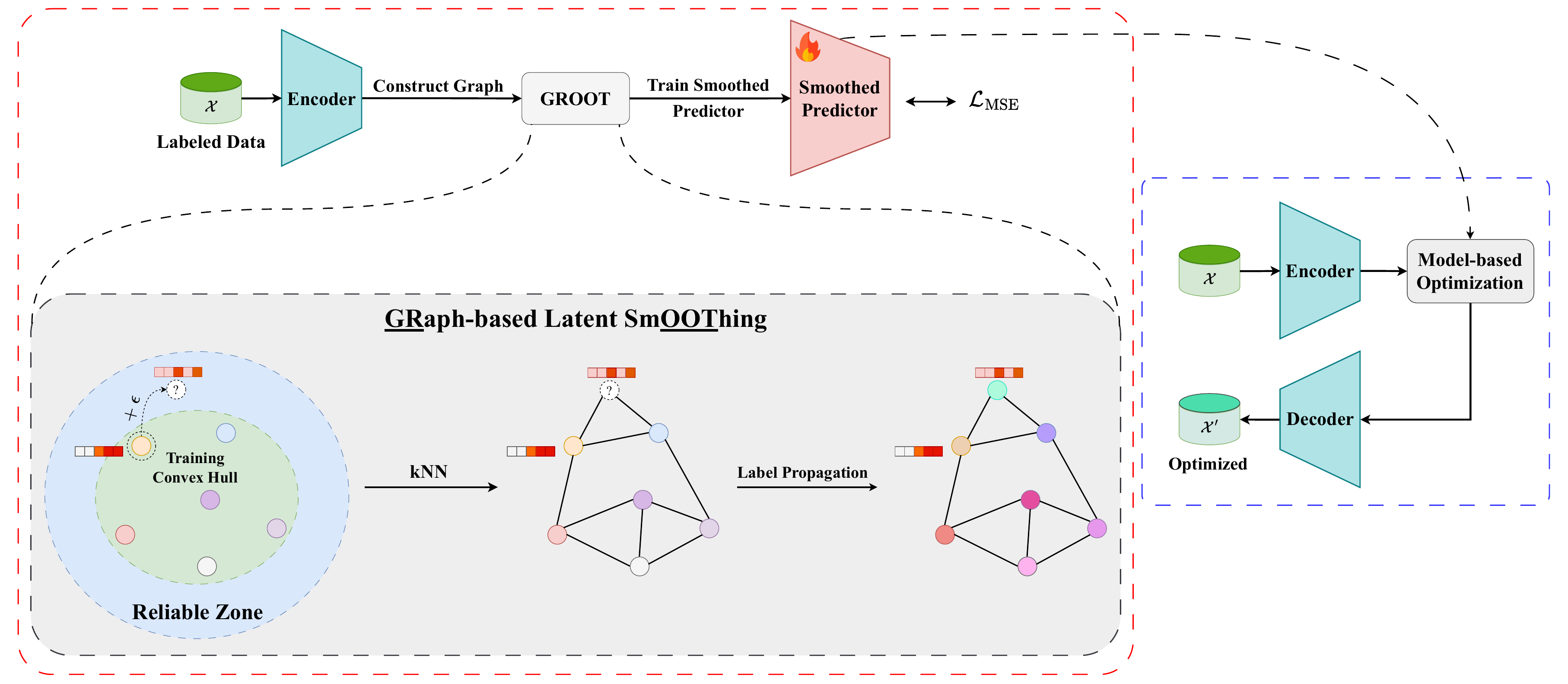}
    \caption{Overall framework of \methodname. After encoding sequences into the latent space, we generate new samples by adding Gaussian noise to existing vectors. These synthetic data lie outside the training set's convex hull but within a reliable zone, as their distances from the hull are below a certain upper bound. We construct a kNN graph and run label propagation to smooth and refine node labels. These nodes and their fitness values are then used to train the surrogate model, which is subsequently employed for optimization.}
    \label{fig:method_overview}
    \Description{Overall framework of \methodname.}
\end{figure*}

We begin this section by formally defining the problem. Next, we detail the construction of the protein latent space and present our proposed latent graph-based smoothing framework, \methodname. \Cref{fig:method_overview} provides a visual overview of our method.

\subsection{Problem Formulation}
We address the challenge of designing proteins to find high-fitness sequences $s$ within the sequence space $\mathcal{A}^L$, where $\mathcal{A}$ represents the amino acid vocabulary (i.e. $|\mathcal{A}| = 20$ since both animal and plant proteins are made up of about 20 common amino acids) and $L$ is the desired sequence length. Our goal is to design sequences that maximize a black-box protein fitness function $\mathcal{O}: \mathcal{A}^L \mapsto \mathbb{R}$, which can only be evaluated through wet-lab experiments:
\begin{equation}
    s^{*} = \underset{s \in \mathcal{A}^L}{\text{argmax}} ~\mathcal{O}(s).
\end{equation}
For in-silico evaluation, given a static dataset of all known sequences and fitness measurements $\mathcal{D}^{*} = \{(s, y) | s \in \mathcal{A}^L, y \in \mathbb{R}\}$, an oracle $\mathcal O_\psi$ parameterized by $\psi$ is trained to minimize the prediction error on $\mathcal{D}^*$. Afterward, $\mathcal{O}_\psi$ is used as an approximator for the black-box function $\mathcal{O}$ to evaluate computational methods that are developed on a training subset $\mathcal{D}$ of $\mathcal{D}^{*}$. In other words, given $\mathcal{D}$, our task is to generate sequences $\hat{s}$ that optimize the fitness approximated by $\mathcal{O}_\psi$. To do this, we train a surrogate model $f$ on $\mathcal{D}$ and use it to guide the search for optimal candidates, which are later evaluated by $\mathcal{O}_\psi$. This setup is similar to work done by \citet{kirjner2024improving, pmlr-v162-jain22a}.

\subsection{Constructing Latent Space of Protein Sequences}\label{sec:vae}
Unsupervised learning has achieved remarkable success in domains like natural language processing \cite{gpt} and computer vision \cite{sam}. This method efficiently learns data representations by identifying underlying patterns without requiring labeled data. The label-free nature of unsupervised learning aligns well with protein design challenges, where experimental fitness evaluations are costly, while unlabeled sequences are abundant. In this work, considering a dataset of unlabeled sequences of a protein family $s \in \mathcal{A} ^ L$ denoted as $\mathcal{D}_u = \{s_i\}_{i=1}^{N_u}$, we train a VAE comprising an encoder $\phi: {\mathcal{A} ^ L} \mapsto \mathbb{R} ^ {d_h}$ and a decoder $\theta: \mathbb{R}^d \mapsto \mathcal{A} ^ L$. Each sequence $s_i$ is encoded into a low-dimensional vector $h_i \in \mathbb{R} ^ {d_h}$. Subsequently, the mean $\mu_i \in \mathbb{R} ^ d$ and log-variance $\log \sigma_i \in \mathbb{R} ^ d$ of the variational posterior approximation are computed from $h_i$ using two feed-forward networks, $\mu_\phi: \mathbb{R} ^ {d_h} \mapsto \mathbb{R} ^ d$ and $\sigma_\phi: \mathbb{R} ^ {d_h} \mapsto \mathbb{R} ^ d$. A latent vector $x_i \in \mathbb{R} ^ d$ is then sampled from the Gaussian distribution $\mathcal{N}(\mu_i, \sigma_i ^ 2)$, and the decoder $\theta$ maps $x_i$ back to the reconstructed protein sequence $\hat{s}_i \in \mathcal{A} ^ L$. The training objective involves the cross-entropy loss $\mathcal{C}(\hat{s}_i, s_i)$ between the ground truth sequence $s_i$ and the generated $\hat s_i$, as well as the Kullback-Leibler (KL) divergence between $\mathcal{N}(\mu_i, \sigma_i ^ 2)$ and $\mathcal{N}(0, I_d)$:
\vspace{-4pt}
\begin{equation}
    \mathcal{L}_{vae} = \frac{1}{N_u}\sum_{i=1} ^ {N_u} \mathcal{C}(\hat{s}_i, s_i) +  \frac{\eta}{N_u} \sum_{i=1} ^ {N_u} D_{\text{KL}} (\mathcal{N}(\mu_i, \sigma_i ^ 2) \| \mathcal{N}(0, I_d)).
    \label{eq:vae_loss}
\end{equation}
Here, $\eta$ is the hyperparameter to control the disentanglement property of the VAE's latent space. A critical distinction must be made between the latent space of protein sequences and the protein fitness landscape. While training a VAE-based model, we do \textit{not} access protein fitness scores, but we learn their continuous representations. Consequently, this VAE-based model is trained on the entirety of available protein family sequences. Subsequently, labeled sequences are encoded by the pretrained VAE into $d$-dimensional latent points, enabling the supervised training of a surrogate model, $\surrogate: \mathbb{R}^d \mapsto \mathbb{R}$, to predict fitness values and approximate the fitness landscape.
\subsubsection{VAE's Architecture} We go into detail regarding the architecture of the VAE used in our study.
\paragraph{\textbf{Encoder}} incorporates a pre-trained ESM-2 \cite{doi:10.1126/science.ade2574} followed by a latent encoder to compute the latent representation $z$. In our study, we leverage the powerful representation of the pre-trained 30-layer ESM-2 by making it the encoder of our model. Given an input sequence $s = \langle s_0, s_1, \cdots, s_L \rangle$, where $s_i \in \mathcal{A}$, ESM-2 computes representations for each token $s_i$ in $s$, resulting in a token-level hidden representation $H = \langle h_0, h_1, \cdots, h_L \rangle, h_i \in \mathbb{R}^{d_h}$. We calculate the global representation $h \in \mathbb{R}^{d_h}$ of $s$ via a weighted sum of its tokens:
\begin{equation}
    h = \sum_{i = 1} ^ L \frac{\omega ^ T \exp{(h_i)}}{\sum_{i = 1} ^ L \omega ^ T \exp{(h_i)}} h_i.
\end{equation}
here, $\omega$ is a learnable global attention vector. Then, two multi-layer perceptrons (MLPs) are used to compute $\mu = \text{MLP}_1(h)$ and $\log \sigma = \text{MLP}_2(h)$, where the latent dimension is $d$. Finally, a latent representation $x \in \mathbb{R}^d$ is sampled from $\mathcal{N}(\mu, \sigma ^ 2)$, which is further proceeded to the decoder to reconstruct the sequence $\hat{s}$. 

\paragraph{\textbf{Decoder}} To decode latent points into sequences, we employ a deep convolutional network as proposed by \citet{Castro2022}, consisting of four one-dimensional convolutional layers. ReLU activations and batch normalization layers are applied between the convolutional layers, except for the final layer.

\subsection{Latent Graph-based Smoothing}\label{sec:smoothing} 

\paragraph{\textbf{Graph Construction}} We need to build a graph $G = (\mathcal{V}, \mathcal{E})$ from the sequences in $\mathcal{D}$ to perform label propagation. \Cref{algo:graph_construction} demonstrates our algorithm to construct a kNN graph from the latent vectors $x$. The graph's nodes are created by sampling $N$ sequences $s \in \mathcal{D}$, and the edges are constructed by $k$-nearest neighbors to the latent embeddings of $x = \phi(s)$, where $\phi$ is the pre-trained encoder. We introduce new nodes to $G$ by interpolating the learned latent $x$ with random noise $\epsilon \in \mathcal{N}(0,I_d)$. We argue that Levenshtein distance is an inadequate metric for computing protein sequence similarity as done in \cite{kirjner2024improving} because marginal amino acid variations can result in substantial fitness disparities \cite{MAYNARDSMITH1970,brookes2022sparsity}. Meanwhile, high-dimensional latent embeddings are effective at capturing implicit patterns within protein sequences. Consequently, Euclidean distance can serve as a suitable metric to quantify sequence similarity for constructing kNN graphs, as proteins with comparable properties (indicated by small fitness score differences) tend to cluster closely in the latent space. \Cref{algo:graph_construction} presents our graph construction pipeline.

\begin{algorithm}[ht]
    \caption{\texttt{CreateGraph}: Latent Graph Construction}
    \noindent 
    \begin{algorithmic}[1]
    \Require training embeddings $\mathcal{V}$, $\#$of graph nodes $N$, constant $\beta$
    \While{$\lvert \mathcal{V} \rvert < N$}
    \State $\overline x \sim \mathcal{U}(\mathcal{V})$
    \State $\epsilon \sim \mathcal{N}(0, I_d)$ \hspace{52pt} \Comment{Sample Gaussian noise $\epsilon$.}
    \State $z \xleftarrow{} \beta * \overline x + (1-\beta) * \epsilon$ \Comment{Interpolate between $ \overline x$ and $\epsilon$.}
    \State $\mathcal{V} \xleftarrow{} \mathcal{V} \cup \{z\}$
    \EndWhile
    \State $\mathcal{E} \xleftarrow{} \bigcup_{x \in \mathcal{V}}\texttt{kNN}(x, \mathcal{V})$ \Comment{Construct edges (\Cref{algo:kNN})}
    \State \Return $G = (\mathcal{V}, \mathcal{E})$
    \end{algorithmic}
    \label{algo:graph_construction}
\end{algorithm}

\paragraph{\textbf{Generating Pseudo-label}} 
We consider a constructed graph G. For each node $v \in \mathcal{V}$, we assign a fitness label $y$, creating a set of fitness labels $Y$. This set $Y$ can be further decomposed into $Y = \{Y_n, Y_u\}$, representing the known and unknown labels, respectively. Known labels $Y_n$ are obtained from the training dataset, while unknown labels $Y_u$ are initialized by 0 and assigned to randomly generated nodes, as shown in line 3 of Algorithm~\ref{algo:smooth}. We then run the following label propagation in $m$ times.  

\begin{equation}
    Y^{\prime} = \alpha D^{-1/2} A D ^ {-1/2} Y + (1-\alpha) Y,
    \label{eq:label_propagation}
\end{equation}
here, $\alpha \in [0, 1]$ is a weighted coefficient and $A$ is a weighted adjacency matrix, defined as: 
\begin{equation}
    A_{ij} = \begin{cases}
        \frac{\gamma}{d(x_i, x_j)} &\text{if } i \neq j, \\
        0 &\text{if } i = j, 
    \end{cases} 
    \label{eq:weighted-adj}
\end{equation}
and $d(\cdot, \cdot)$ is the Euclidean distance, $\gamma$ is a controllable factor. In summary, \Cref{algo:smooth} details our smoothing strategy by label propagation.

\begin{algorithm}[ht]
    \caption{\texttt{Smooth}: Latent Graph-based Smoothing}
    \noindent 
    \begin{algorithmic}[1]
    \Require graph $G = (\mathcal V, \mathcal E)$, label propagation layers $m$
    \State $A \leftarrow \texttt{WeightedAdjacencyMatrix}(\mathcal E)$ \Comment{\Cref{eq:weighted-adj}}
    \State $D \leftarrow \texttt{DegreeMatrix}(\mathcal V)$ \Comment{Compute degree matrix.}
    \State ${Y} \leftarrow {[0 \text{ if } \texttt{is\_synthetic} \text{ else } y_i]_{i=1}^N}$
    \For{$i = 0,\ldots, m - 1$}
        \State $Y \leftarrow \texttt{LabelPropagation}(A,D,Y)$ \Comment{\Cref{eq:label_propagation}}
    \EndFor
    \State \Return $Y$
    \end{algorithmic}
    \label{algo:smooth}
\end{algorithm}

\subsection{Theoretical Justification}
This section delves into the theoretical foundations of our proposed method. We begin by discussing the concepts of interpolation and extrapolation within high-dimensional latent embedding spaces. These concepts ensure the correctness of generating additional pseudo-samples within the latent space of a VAE-based model. Subsequently, leveraging the definition of convex hulls, we establish an upper bound for the expected distance between the generated nodes and the collection of training latent embeddings. This bound informs the intuition behind our kNN graph construction, detailed in ~\Cref{algo:graph_construction}. Consequently, employing label propagation within this bounded distance emerges as a rational strategy for assigning pseudo-labels to the newly generated nodes.

We first list the definitions of a convex hull, interpolation, and extrapolation as follows:
\begin{definition}
    Let $\mathbb{X} = \{x_1, \dots, x_N \}$ be a set of $N$ points in $\mathbb{R}^d$, a convex hull of $\mathbb{X}$ is defined as:
    \begin{equation*}
        \conv \triangleq \bigg\{\sum_{i=1}^N \lambda_i x_i | \lambda_i \ge 0 \text{ and } \sum_{i=1}^N \lambda_i = 1\bigg\}.
    \end{equation*}
\end{definition}

\begin{definition}[\citet{extrapolate}] \label{def:interpolation}
    Interpolation occurs for a sample $x$ whenever this sample belongs to $\conv$, if not, extrapolation occurs.
\end{definition}

\paragraph{\textbf{Assumptions}} Given a set of $N$ training protein sequences $\mathbb{S} = \{s_1, s_2, \dots, s_N\}$, we use the encoder $\phi$ of the pretrained VAE to compute their latent embeddings, resulting in a set of latent vectors $\mathbb{X} = \{x_1, x_2, \dots, x_N\}$, where $x_i = \phi(s_i)$ (see \Cref{sec:vae}). In this paper, we consider the scenario wherein we have limited access to experimental data, and we train a VAE-based model to represent continuous $d$-dimensional latent vectors of those available sequences. This, therefore, gives rise to two mild assumptions in our work. 
\begin{itemize}
    \item \textbf{Assumption 1}: $N$ is significantly smaller than the exponential in $d$, where $N$ represents the number of data samples.
    \item \textbf{Assumption 2}: Each latent vector $x$ is sampled from $\mathcal{N}(0, I_d)$. This is achieved by regularizing the model with $KL$ divergence loss as shown in the \Cref{eq:vae_loss}.
\end{itemize}

We assess \methodname's extrapolation capabilities by computing the probability of its generated samples falling outside the convex hull of the training data within the latent space. Subsequently, we derive an upper bound for these reliable extrapolation regions. These findings are formalized in the following propositions.

\begin{proposition}
     Let $\mathbb{X} = \{x_i \in \mathbb{R}^d\}_{i=1}^N$ be a set of $N$ i.i.d $d$-dimensional samples from $\mathcal{N}(0,I_d)$. Assume that $N \ll \exp \big(\frac{d}{2(C^2_{\beta}+2)}\big)$ and let $z = \beta * \overline x + (1-\beta) * \epsilon$ for some $\overline x \in \mathbb{X}$, $\beta \in (0,1)$, $C_{\beta} = \frac{1+\beta}{1-\beta}$ and $\epsilon \sim \mathcal{N}(0,I_d)$, then
     \begin{equation*}
         \lim_{d \rightarrow \infty} \mathbf{P}(z \notin \conv) = 1.
     \end{equation*}
     \label{prop:prob_limit}
\end{proposition}
We leave the proof of the presented proposition in \Cref{app:proof_of_prob_limit}. \Cref{prop:prob_limit} elucidates that in a limited data scenario (i.e. $N \ll \exp\left(\frac{d}{2(C^2_{\beta} + 2)}\right)$), the probability for a synthetic latent node $z$ lying outside the convex hull of training set $\mathbb{X}$ goes to 1 as the latent dimension $d$ grows. Based on \Cref{def:interpolation}, this theoretical result guarantees the correctness of our formula shown in line 4 of \Cref{algo:graph_construction} in expanding the latent space of protein sequences. However, when generated nodes are located far from the training convex hull, their embedding vectors differ significantly from those in the training set. This disparity hinders the identification of meaningful similarities and can result in unreliable fitness scores determined through label propagation. This brings us to \Cref{prop:distance}.
 
\begin{proposition}\label{prop:distance}
   Let $\mathbb{X} = \{x_i \in \mathbb{R}^d\}_{i=1}^N$ be a set of $N$ i.i.d $d$-dimensional samples from $\mathcal{N}(0,I_d)$. For any $\overline{x} \in \mathbb{X}$, we define $z = \beta * \overline x + (1-\beta) * \epsilon$ with $\epsilon \sim \mathcal{N}(0,I_d), \beta \in (0,1)$, the following holds:
   \begin{equation*}
       \mathbb{E}[D(z,\conv)] < 2(1-\beta)\sqrt{d},
   \end{equation*}  
   where $D(z, \conv)) \triangleq \inf_{x \in \conv} d(z, x)$ is the distance from a point $z \in \mathbb{R}^d$ to $\conv$, and $d(\cdot, \cdot)$ is the Euclidean distance. 
\end{proposition}
The proof of this proposition can be found in \Cref{app:prop_distance}. \Cref{prop:distance} allows us to quantify the expected distance from a randomly generated node based on our formula to the available training set. The derived upper bound increases linearly w.r.t square root of the latent dimension $d$ with a rate of $2(1-\beta)$. It is worth noting that $\beta$ is used to control the exploration rate of our algorithm. As $\beta \rightarrow 1$, the generated node $z$ closely resembles a training node $\overline x$. Conversely, a lower $\beta$ introduces more noise into the latent vectors, resulting in samples that are further from the source nodes. This upper bound ensures that synthetic nodes remain within a controllable \textit{"reliable zone"}, making label propagation a sensible choice, as the label values of these synthetic nodes cannot deviate too far from the overall distribution of existing nodes.

In summary, Propositions \ref{prop:prob_limit} and \ref{prop:distance} offer theoretical underpinnings for our latent-based smoothing approach. By mapping discrete sequences to a continuous $d$-dimensional latent space, we establish a quantitative relationship between newly added and original training nodes, governed by parameters $d$ and $\beta$. This quantitative connection enhances the interpretability of our method. 

\subsection{Model-based Optimization Algorithm} 
Latent space model-based optimization (MBO) employs surrogate models, $f_\Phi: \mathbb{R}^d \mapsto \mathbb{R}$, to efficiently guide the search for optimal values that maximize expensive black-box functions. The effectiveness of MBO depends on the accuracy of the surrogate model as poorly trained surrogates can hinder the optimization process by providing misleading information. As shown in \Cref{algo:groot}, lines 1, 4, and 5 encompass the standard MBO process of encoding training vectors into latent space and training a surrogate model to predict fitness scores Y based on these embeddings. Meanwhile, lines 2 and 3 are two additional steps introduced by \method before surrogate training. \method is agnostic to domains as we can use any encoding method to compute latent embeddings of target domains. Moreover, \method can be used in tandem with any surrogate models as it only accesses the data representations. These two properties make \method a versatile approach for optimization tasks in a wide range of domains. 
\begin{algorithm}[h]
    \caption{\texttt{\method}}
    \noindent
    \begin{algorithmic}[1]
    \Require training dataset $\mathcal{D}$, pre-trained encoder $\phi$, $\#$ of graph nodes $N$, constant $\beta$, label propagation layers $m$
    \State $\mathcal{V} \leftarrow \{x_i = \phi(s_i), s_i \in \mathcal{D}\}_{i=1}^n$ \Comment{Compute latent embeddings.}
    \State $\mathcal V', \mathcal E \leftarrow \texttt{CreateGraph}(\mathcal V, N, \beta)$ \Comment{\Cref{algo:graph_construction}}
    \State $\hat Y \leftarrow \texttt{Smooth}\big((\mathcal V', \mathcal E), L \big)$ \Comment{\Cref{algo:smooth}}
    \State $\Phi \leftarrow \arg\min_{\Phi} \mathbb{E}_{(x,y) \sim (\mathcal V', \hat Y)} \big[ (y - f_\Phi(x))^2 \big]$ \Comment{Train surrogate model.}
    \State $X^*, Y^* \leftarrow \texttt{MBO}(\Phi, \mathcal V')$ \Comment{Model-based optimization.}
    \State \Return $X^*, Y^*$
    \end{algorithmic}
    \label{algo:groot}
\end{algorithm}
\section{Experiments} \label{sec:experiments}




We demonstrate the benefits of our proposed \method on two protein tasks with varying levels of label scarcity. Furthermore, since obtaining the ground-truth fitness of generated sequences is costly, we further evaluate our method on Design-Bench \cite{pmlr-v162-trabucco22a}. These benchmarks have exact oracles, providing a better validation of our method's performance on real-world datasets.

\paragraph{\textbf{Design-Bench}}
Due to space constraints, we leave the presentation of tasks, baselines, evaluation metrics, and implementation details in \Cref{app:design-bench}. \Cref{tab:db} shows the performance of various methods across three Design-Bench tasks and their mean scores. While most methods perform well, \method achieves the highest average score in maximum performance and is slightly behind the SOTA baseline in median and mean benchmarks. Since these tasks are evaluated by an exact oracle and our performance is competitive with other SOTA methods, this confirms that our approach is effective on real-world data, enhancing the reliability of our work.

\begin{table}[t]
    \centering
    \caption{Comparison of \method and the baselines on 3 Design-Bench tasks. \textbf{Bold} results indicate the best value, and \underline{underlined} results indicate the second-best value. The standard deviation of 3 runs with different random seeds is indicated in parentheses.}
    \label{tab:db}
    \begin{tabular}{cccccc}
        \toprule
         & \multirow{2}{*}{Method} & \multirow{2}{*}{D'Kitty} & \multirow{2}{*}{Ant} & \multirow{2}{*}{TF Bind 8} & Mean \\
         & & & & & Score $\uparrow$ \\
         \midrule
         \multirow{5}{*}{Median} & MINs & \underline{0.86 \scriptsize{(0.01)}} & 0.49 \scriptsize{(0.15)} & 0.42 \scriptsize{(0.02)} & 0.59 \\
         & BONET & 0.85 \scriptsize{(0.01)} & {0.60 \scriptsize{(0.12)}} & 0.44 \scriptsize{(0.00)} & 0.63 \\
         & BDI & 0.59 \scriptsize{(0.02)} & 0.40 \scriptsize{(0.02)} & \textbf{0.54 \scriptsize{(0.03)}} & 0.51 \\
         & ExPT & \textbf{0.90 \scriptsize{(0.01)}} & \textbf{0.71 \scriptsize{(0.02)}} & 0.47 \scriptsize{(0.01)} & \textbf{0.69} \\
         & \methodname & 0.82 \scriptsize{(0.01)} & \underline{0.62 \scriptsize{(0.01)}} & \underline{0.52 \scriptsize{(0.01)}} & \underline{0.65} \\
         \midrule
         \multirow{5}{*}{Max} & MINs & 0.93 \scriptsize{(0.01)} & 0.89 \scriptsize{(0.01)} & 0.81 \scriptsize{(0.03)} & 0.88 \\
         & BONET & 0.91 \scriptsize{(0.01)} & 0.84 \scriptsize{(0.04)} & 0.69 \scriptsize{(0.15)} & 0.81 \\
         & BDI & 0.92 \scriptsize{(0.01)} & 0.81 \scriptsize{(0.09)} & 0.91 \scriptsize{(0.07)} & 0.88 \\
         & ExPT & \underline{0.97 \scriptsize{(0.01)}} & \textbf{0.97 \scriptsize{(0.00)}} & \underline{0.93 \scriptsize{(0.04)}} & \underline{0.96} \\
         & \methodname & \textbf{0.98 \scriptsize{(0.01)}} & \textbf{0.97 \scriptsize{(0.00)}} & \textbf{0.98 \scriptsize{(0.02)}} & \textbf{0.98} \\
         \midrule
         \multirow{5}{*}{Mean} & MINs & 0.62 \scriptsize{(0.03)} & 0.01 \scriptsize{(0.01)} & 0.42 \scriptsize{(0.03)} & 0.35 \\
         & BONET & \underline{0.84 \scriptsize{(0.02)}} & {0.58 \scriptsize{(0.02)}} & 0.45 \scriptsize{(0.01)} & 0.62 \\
         & BDI & 0.57 \scriptsize{(0.03)} & 0.39 \scriptsize{(0.01)} & \underline{0.54 \scriptsize{(0.03)}} & 0.50 \\
         & ExPT & \textbf{0.87 \scriptsize{(0.02)}} & \textbf{0.64 \scriptsize{(0.03)}} & 0.48 \scriptsize{(0.01)} & \textbf{0.66} \\
         & \methodname & 0.78 \scriptsize{(0.03)} & \underline{0.60 \scriptsize{(0.01)}} & \textbf{0.55 \scriptsize{(0.02)}} & \underline{0.64} \\
         \bottomrule
    \end{tabular}
\end{table}

\subsection{Experiment Setup}\label{sec:exp-setup}

\paragraph{\textbf{Datasets and Oracles}} Following \citet{kirjner2024improving}, we evaluate our method on two proteins, GFP \cite{sarkisyan2016local} and AAV \cite{bryant2021deep}. The length $L$ is 237 for GFP and 28 for the functional segment of AAV. The full GFP dataset $\mathcal{D^*}$ comprises 54,025 mutant sequences with associated log-fluorescence intensity, while the AAV dataset contains 44,156 sequences linked to their ability to package a DNA payload. The fitness values in both datasets are min-max normalized for evaluation but remain unchanged during training and inference. To demonstrate the advantages of our smoothing method, we use the \textit{harder1}, \textit{harder2}, and \textit{harder3} level benchmarks proposed by \citet{kirjner2024improving} to sample training datasets $\mathcal{D}$ for each task, simulating scenarios with scarce labeled data\footnote{Results of other difficulties are presented in the \Cref{app:protein}}. \Cref{tab:stats} provides statistics for each level benchmark. All methods, including baselines, start optimization using the entire training set $\mathcal{D}$ and are evaluated on the best 128 generated sequences, with approximated fitness predicted by oracles provided by \citet{kirjner2024improving}. It is important to note that oracles do not participate in the optimization process and are only used as \textit{in-silico} evaluators to validate final proposed designs of each method.

\begin{table}[t]
\centering
\caption{Statistic of benchmarks.}{
    \label{tab:stats}
    \begin{tabular}{lccccc}
        \toprule
        \multirow{2}{*}{Task} & \multirow{2}{*}{Difficulty} & Fitness & Mutational & Best & \multirow{2}{*}{$|\mathcal D|$} \\
        & & Range ($\%$) & Gap & Fitness & \\
        \midrule
        \multirow{3}{*}{AAV} & Harder1 & $< 30$th & $13$ & $0.33$ & $1157$ \\ 
        & Harder2 & $< 20$th & $13$ & $0.29$ & $920$ \\ 
        & Harder3 & $< 10$th & $13$ & $0.24$ & $476$ \\
        \midrule
        \multirow{3}{*}{GFP} & Harder1 & $< 30$th & $8$ & $0.10$ & $1129$ \\ 
        & Harder2 & $< 20$th & $8$ & $0.01$ & $792$ \\ 
        & Harder3 & $< 10$th & $8$ & $0.01$ & $397$ \\
        \bottomrule
    \end{tabular}
}
\end{table}

\begin{table*}[t]
    \centering
    \caption{\textbf{AAV and GFP optimization results} for \method and baseline methods. The standard deviation of 5 runs with different random seeds is indicated in parentheses.}
    \label{tab:numeric}
    \begin{tabular}{lccc|ccc|ccc}
    \toprule
         & \multicolumn{3}{c}{AAV \textit{harder1} task} & \multicolumn{3}{c}{AAV \textit{harder2} task} & \multicolumn{3}{c}{AAV \textit{harder3} task} \\ 
         \cmidrule(lr){2-4} \cmidrule(lr){5-7} \cmidrule(lr){8-10}
         Method & Fitness $\uparrow$ & Diversity & Novelty & Fitness $\uparrow$ & Diversity & Novelty & Fitness $\uparrow$ & Diversity & Novelty \\
    \midrule
        AdaLead & {0.38 (0.0)} & 5.5 (0.5) & 7.0 (0.7) & 0.43 (0.0) & 4.2 (0.7) & 7.8 (0.8) & 0.37 (0.0) & 6.22 (0.9) & 8.0 (1.2) \\
        CbAS & 0.02 (0.0) & 22.9 (0.1) & 18.5 (0.5) & 0.01 (0.0) & 23.2 (0.1) & 19.3 (0.4) & 0.01 (0.0) & 23.2 (0.1) & 19.3 (0.4) \\
        BO & 0.00 (0.0) & 20.4 (0.3) & 21.8 (0.4) & 0.01 (0.0) & 20.4 (0.0) & 22.0 (0.0) & 0.01 (0.0) & 20.6 (0.3) & 22.0 (0.0) \\
        GFN-AL & 0.00 (0.0) & 15.4 (6.2)& 21.6 (0.5) & 0.00 (0.0) & 8.1 (3.5) & 21.6 (1.0) & 0.00 (0.0) & 7.6 (0.8) & 22.6 (1.4) \\
        PEX & 0.23 (0.0) & 6.4 (0.5) & 3.8 (0.7) & 0.30 (0.0) & 7.8 (0.4) & 5.0 (0.0) & 0.26 (0.0) & 7.3 (0.7) & 4.4 (0.5) \\
        GGS & 0.30 (0.0) & 13.6 (0.2) & 14.5 (0.3) & 0.27 (0.0) & 16.0 (0.0) & 19.4 (0.0) & 0.38 (0.0) & 7.0 (0.1) & 9.6 (0.1) \\
        ReLSO & 0.15 (0.0) & 20.9 (0.0) & 13.0 (0.0) & 0.17 (0.0) & 20.3 (0.0) & 13.0 (0.0) & 0.22 (0.0) & 17.8 (0.0) & 11.0 (0.0) \\
        S-ReLSO & 0.24 (0.0) & 11.5 (0.0) & 13.0 (0.0) & 0.28 (0.0) & 16.4 (0.0) & 6.5 (0.0) & 0.27 (0.0) & 17.7 (0.0) & 11.0 (0.0) \\
        \textbf{\method} & \textbf{0.46 (0.1)} & 9.8 (1.6) & 12.2 (0.5) & \textbf{0.45 (0.0)} & 9.9 (0.8) & 13.0 (0.0) & \textbf{0.42 (0.1)} & 11.0 (2.0) & 13.0 (0.0) \\
    \toprule
        & \multicolumn{3}{c}{GFP \textit{harder1} task} & \multicolumn{3}{c}{GFP \textit{harder2} task} & \multicolumn{3}{c}{GFP \textit{harder3} task} \\ 
         \cmidrule(lr){2-4} \cmidrule(lr){5-7} \cmidrule(lr){8-10}
         Method & Fitness $\uparrow$ & Diversity & Novelty & Fitness $\uparrow$ & Diversity & Novelty & Fitness $\uparrow$ & Diversity & Novelty \\
    \midrule
        AdaLead & 0.39 (0.0) & 8.4 (3.2) & 9.0 (1.2) & 0.4 (0.0) & 7.3 (2.8) & 9.8 (0.4) & 0.42 (0.0) & 6.4 (2.3) & 9.0 (1.2) \\
        CbAS & -0.08 (0.0) & 172.2 (35.7) & 201.5 (1.5) & -0.09 (0.0) & 158.4 (34.8) & 202.0 (0.7) & -0.08 (0.0) & 186.4 (33.4) & 201.5 (0.9)\\
        BO & -0.08 (0.1) & 58.9 (1.9) & 192.3 (11.3) & -0.04 (0.1) & 57.1 (1.7) & 192.3 (11.3) & -0.07 (0.1) & 57.8 (2.2) & 177.9 (41.2) \\
        GFN-AL & 0.21 (0.1) & 74.3 (55.3) & 219.2 (3.3) & 0.14 (0.2) & 27.0 (9.5) & 223.5 (2.4) & 0.21 (0.0) & 37.5 (21.7) & 219.8 (4.3) \\
        PEX & 0.13 (0.0) & 12.6 (1.2) & 7.1 (1.1) & 0.17 (0.0) & 12.6 (1.2) & 7.1 (1.1) & 0.19 (0.0) & 12.2 (1.1) & 7.8 (1.7) \\
        GGS & 0.67 (0.0) & 4.7 (0.2) & 9.1 (0.1) & 0.60 (0.0) & 5.4 (0.2) & 9.8 (0.1) & 0.00 (0.0) & 15.7 (0.4) & 19.0 (2.2) \\
        ReLSO & \textit{0.94 (0.0)}$^\dagger$ & 0.0 (0.0) & 8.0 (0.0) & \textit{0.94 (0.0)}$^\dagger$ & 0.0 (0.0) & 8.0 (0.0) & \textit{0.94 (0.0)}$^\dagger$ & 0.0 (0.0) & 8.0 (0.0) \\
        S-ReLSO & \textit{0.94 (0.0)}$^\dagger$ & 0.0 (0.0) & 8.0 (0.0) & \textit{0.94 (0.0)}$^\dagger$ & 0.0 (0.0) & 8.0 (0.0) & \textit{0.94 (0.0)}$^\dagger$ & 0.0 (0.0) & 8.0 (0.0) \\
        \textbf{\method} & \textbf{0.88 (0.0)} & 3.0 (0.2) & 7.0 (0.0) & \textbf{0.87 (0.0)} & 3.0 (0.1) & 7.5 (0.5) & \textbf{0.62 (0.2)} & 7.6 (1.5) & 8.6 (1.5) \\
    \bottomrule
    \multicolumn{10}{l}{$\dagger$ indicates that the generated population has collapsed (i.e., producing only a single sequence).} \\
    \end{tabular}
    \setlength{\tabcolsep}{3pt}
\end{table*}

\paragraph{\textbf{Baselines}} We evaluate our method against several representative baselines using the open-source toolkit FLEXS \cite{DBLP:journals/corr/abs-2010-02141}: AdaLead \cite{DBLP:journals/corr/abs-2010-02141}, CbAS \cite{brookes2019conditioning}, and Bayesian Optimization (BO) \cite{wilson2017reparameterization}. In addition to these algorithms, we also benchmark against some most recent methods: GFN-AL \cite{pmlr-v162-jain22a}, ReLSO \cite{Castro2022}, PEX \cite{pmlr-v162-ren22a}, and GGS \cite{kirjner2024improving}, utilizing and modifying the code provided by their respective authors. Furthermore, we assess ReLSO enhanced with our smoothing strategy for better optimization, termed S-ReLSO.

\paragraph{\textbf{Evaluation Metrics}} We use three metrics proposed by \citet{pmlr-v162-jain22a}: fitness, diversity, and novelty. Let the optimized sequences $\mathcal G^* = \{ g_1^*, \ldots, g_K^* \}$. Fitness is defined as the median of the evaluated fitness of $K = 128$ proposed designs. Diversity is the median of the distances between every pair of sequences in $\mathcal G^*$. Novelty is defined as the median of the distances between every generated sequences to the training set. Mathematical definitions of these metrics are defined in the \Cref{app:metrics}.

\paragraph{\textbf{Implementation Details}} For each task, we use the full dataset $\mathcal D^*$ to train the VAE, whose architecture has been described in \Cref{sec:vae}. We utilize the pretrained checkpoint \texttt{esm2\_t30\_150M\_UR50D}\footnote{\url{https://huggingface.co/facebook/esm2_t30_150M_UR50D}} of the ESM-2 model, fine-tuning only the last layer while freezing the rest. The latent representation space dimension is set to $d = 320$. For the latent graph-based smoothing, we set the number of nodes to $N=20,000$, label propagation's coefficient to $\alpha=0.2$, controller factor to $\gamma = 1.0$, the number of propagation layers to $N_{\text{layers}}=1$, and the number of neighbors to $k=8$. It is important to note that these hyperparameters are not finely tuned. The hyperparameter tuning process is described in the \Cref{app:method-hparams}. After the smoothing process, the features and refined labels are inputted into a surrogate model, specifically a shallow 2-layer multi-layers perceptron (MLP) with a dropout rate of $0.2$ in the first layer, to train a smoothed surrogate model. Despite its simplicity, we consider MLP a suitable choice due to its proven effectiveness in previous studies \cite{huang2021combining}. For model-based optimization, we select two gradient-based algorithms to exploit the smoothness of the surrogate model: Gradient Ascent (GA) with a learning rate of $0.005$ for $400$ iterations and Limited-memory Broyden-Fletcher-Goldfarb-Shanno (L-BFGS) \cite{Liu1989} for a maximum of $6$ iterations. While various optimization algorithms are available, we choose these gradient-based methods based on the premise that they can effectively leverage the smoothness of the surrogate model. It is crucial to note that \textit{our primary focus is not on evaluating multiple optimization methods, but rather on demonstrating the effectiveness of the smoothing strategy}. We empirically evaluate these two algorithms and report the best results.

\subsection{Numerical Results}
\label{sec:numerical-results}
We report the mean and standard deviation of the evaluation metrics over five runs with different random seeds in \Cref{tab:numeric}. Firstly, compared to the best fitness of the training data (\Cref{tab:stats}), our method successfully optimizes all tasks. For AAV tasks, our method achieves SOTA results across all difficulty levels. Notably, although our method only slightly exceeds AdaLead in the \textit{harder2} task, \method generates sequences whose novelty falls within the mutation gap range, indicating appropriate extrapolation and a higher likelihood of producing synthesizable proteins. Additionally, applying our smoothing strategy to ReLSO improves the model's performance by $60\%$ in the \textit{harder1}, $64.7\%$ in the \textit{harder2}, and $22.7\%$ in the \textit{harder3} tasks. For GFP tasks, \method outperforms all baseline methods across all difficulties, especially in the \textit{harder3} difficulty, where GGS failed to optimize properly. It is crucial to mention that ReLSO failed to generate diverse sequences in GFP tasks, as indicated by zero diversity across five different runs. Overall, we observe that \method achieves the highest fitness while maintaining respectable diversity and novelty.

\subsection{Analysis}
\paragraph{\textbf{Impact of smoothing on extrapolation and general performance}} 
For each benchmark, we assess the impact of smoothing on extrapolation capabilities by measuring the Mean Absolute Error (MAE) of the surrogate model on the benchmark's training and holdout datasets relative to the experimental ground-truth. \Cref{tab:smooth-mae} shows the benefits of smoothing on extrapolation to held out ground-truth experimental data. Our results demonstrate that smoothing reduces MAE for both the training and holdout sets. This reduction occurs because our smoothing strategy acts as a data augmentation technique in the latent space, enhancing the robustness of the supervised model. We also observe that the MAE for both sets increases gradually as the task becomes more difficult. We hypothesize that this is due to the smaller training set size in more difficult tasks, requiring a greater number of new samples to construct a graph. While the labels of these new samples are not exact, they help to smooth the latent landscape, resulting in a smoother but not perfectly accurate surrogate model, which in turn increases the MAE.

Additionally, \Cref{tab:smooth-fitness} demonstrates how a smoothed model dramatically outperforms its unsmoothed counterpart across all tasks. For AAV, the results show up to a threefold improvement compared to the unsmoothed surrogate model. For GFP, without smoothing, the optimization algorithm struggles to find a maximum point, resulting in suboptimal performance. With our smoothing technique, the optimization algorithm can optimize properly. Our findings indicate that our smoothing strategy enables gradient-based optimization methods to better navigate the search space and identify superior maxima. However, we also observe that applying smoothing decreases diversity. This can be explained by the optimization process converging, causing the population to concentrate in certain high-fitness regions. In contrast, without proper optimization, the population diverges and spreads throughout the search space.

\begin{table}[t]
\centering
\caption{Smoothing improves extrapolation on every tasks. The mean of 5 different runs is reported.}{
    \label{tab:smooth-mae}
    \begin{tabular}{lcccc}
        \toprule
        \multirow{2}{*}{Task} & \multirow{2}{*}{Difficulty} & \multirow{2}{*}{Smoothed} & Train & Holdout \\
        & & & MAE $\downarrow$ & MAE $\downarrow$ \\
        \midrule
        \multirow{6}{*}{AAV} & \multirow{2}{*}{Harder1} & No & 4.94 & 8.93 \\
        & & Yes & \textbf{1.02} & \textbf{5.78} \\ \cmidrule{2-5}
        & \multirow{2}{*}{Harder2} & No & 4.67 & 8.91 \\
        & & Yes & \textbf{1.10} & \textbf{6.11} \\ \cmidrule{2-5}
        & \multirow{2}{*}{Harder3} & No & 4.13 & 8.87 \\
        & & Yes & \textbf{1.44} & \textbf{7.30} \\
        \midrule
        \multirow{6}{*}{GFP} & \multirow{2}{*}{Harder1} & No & 1.39 & 2.81 \\
        & & Yes & \textbf{0.22} & \textbf{1.71} \\ \cmidrule{2-5}
        & \multirow{2}{*}{Harder2} & No & 1.34 & 2.81 \\
        & & Yes & \textbf{0.31} & \textbf{1.85} \\ \cmidrule{2-5}
        & \multirow{2}{*}{Harder3} & No & 1.33 & 2.80 \\
        & & Yes & \textbf{0.51} & \textbf{2.09} \\
        \bottomrule
    \end{tabular}
}
\end{table}

\begin{table}[t]
\centering
\caption{Smoothing improves performance on every tasks. The mean of 5 different runs is reported.}{
    \label{tab:smooth-fitness}
    \begin{tabular}{lccccc}
        \toprule
        Task & Difficulty & Smoothed & Fitness $\uparrow$ & Diversity & Novelty \\
        \midrule
        \multirow{6}{*}{AAV} & \multirow{2}{*}{Harder1} & No & 0.12 & 20.0 & 10.0 \\
        & & Yes & \textbf{0.46} & {9.8} & {12.2} \\ \cmidrule{2-6}
        & \multirow{2}{*}{Harder2} & No & 0.11 & 20.0 & 9.6 \\
        & & Yes & \textbf{0.45} & {9.9} & {13.0} \\ \cmidrule{2-6}
        & \multirow{2}{*}{Harder3} & No & 0.12 & 20.1 & 10.0 \\
        & & Yes & \textbf{0.42} & {11.0} & {13.0} \\
        \midrule
        \multirow{6}{*}{GFP} & \multirow{2}{*}{Harder1} & No & -0.12 & 71.0 & 42.2 \\
        & & Yes & \textbf{0.88} & {3.0} & {7.0} \\ \cmidrule{2-6}
        & \multirow{2}{*}{Harder2} & No & -0.18 & 69.5 & 41.1 \\
        & & Yes & \textbf{0.87} & {3.0} & {7.5} \\ \cmidrule{2-6}
        & \multirow{2}{*}{Harder3} & No & -0.17 & 64.0 & 37.0 \\
        & & Yes & \textbf{0.62} & {7.6} & {8.6} \\
        \bottomrule
    \end{tabular}
}
\end{table}

\begin{figure}[ht]
    \centering
    \includegraphics[width=1.0\linewidth]{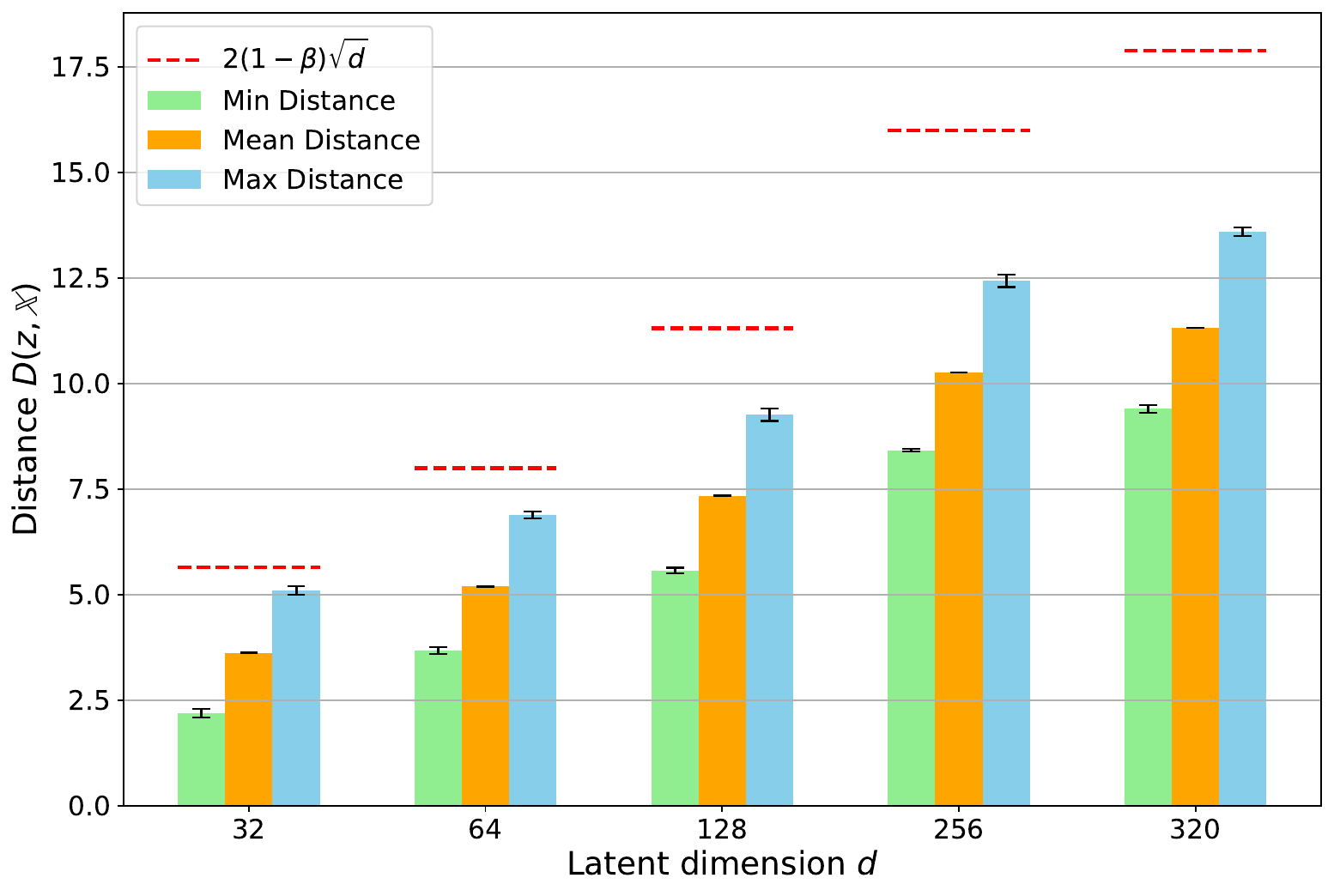}
    \caption{Distance from generated nodes outside the convex hull to the set of training nodes. The mean and standard deviation over 5 different runs are reported. We set $\boldsymbol{\beta=0.5}$ to maintain a constant upper bound on the distance.}
    \Description{Distance from generated nodes outside the convex hull to the set of training nodes.}
    \label{fig:proposition}
\end{figure}

\paragraph{\textbf{Empirical analysis of our propositions}} In this section, we directly validate our proposed \Cref{prop:prob_limit,prop:distance} through experiments. Firstly, we train multiple VAEs with the same architecture described in \Cref{sec:vae}, modifying only the latent dimension size $d$. For each VAE version, we construct a graph and apply the smoothing technique as described in \Cref{sec:smoothing} to generate new nodes. Secondly, to determine whether the generated nodes lie within the convex hull, we check if each point can be expressed as a convex combination of the training points. This is formulated as a linear programming problem and can be easily solved using existing libraries\footnote{\url{https://docs.scipy.org/doc/scipy/reference/generated/scipy.optimize.linprog.html}}. After identifying the number of nodes lying outside the convex hull, we calculate the Euclidean distance $D(z, \mathbb{X})$ between the outside nodes and the set of training nodes.

We validate our propositions on the GFP harder3 task, setting $\beta = 0.5$ consistently across all runs to maintain a constant upper bound. Our experiments reveal that, regardless of dimensional size, 100\% of the synthetic nodes lie outside the convex hull. \Cref{fig:proposition} demonstrates distance calculations between the outside nodes and the set of training nodes. It is evident that although the distance increases with the latent dimension, none of the generated nodes in any VAE versions exceed the proven upper bound. This demonstrates that our smoothing technique can expand into extrapolation regions while remaining within a suitable range, avoiding distances too far from the training set that could introduce errors in surrogate model predictions.

\paragraph{\textbf{Explore the limits of smoothing strategy}} In this section, we investigate the smoothing capabilities of our method by testing its limits. We use the GFP \textit{harder3} task and the AAV \textit{harder3} task as examined cases, further subsampling the dataset with ratios $r \in \{ 0.05, 0.1, 0.2, 0.5, 0.7, 1.0 \}$ in two different ways, namely: \textit{random} and \textit{lowest}. In the \textit{random} setting, data points are randomly subsampled with the given ratio $r$, while in the \textit{lowest} setting, we use the fraction $r$ of the dataset with the lowest fitness scores.

\begin{figure}
    \centering
    \includegraphics[width=\linewidth]{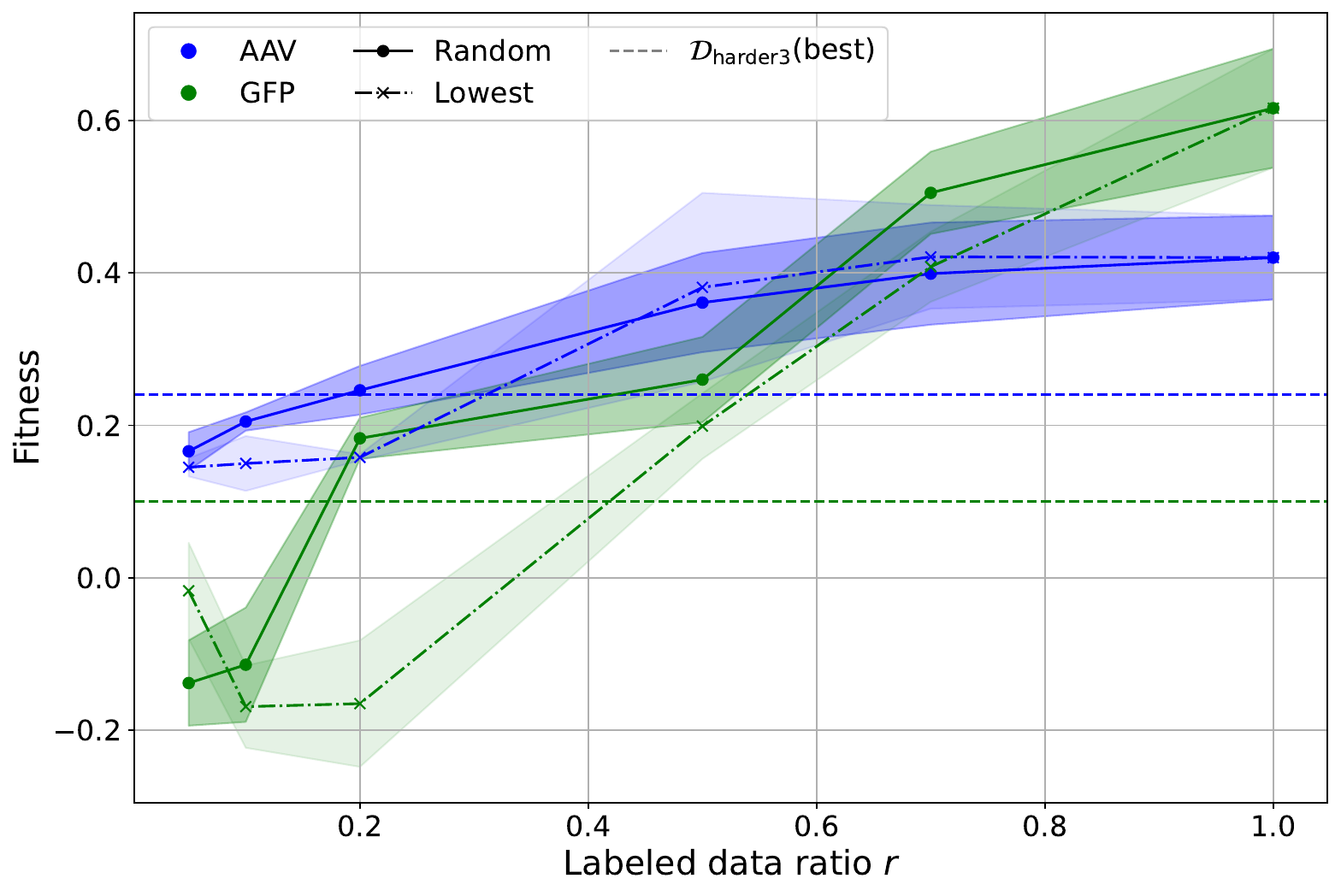}
    \caption{The performance of \method on AAV and GFP \textit{harder3} tasks when we vary the labeled data ratio $r$. The mean and standard deviation over 5 different runs are reported.}
    \Description{Performance when varying the training data ratio $r$.}
    \label{fig:limit}
\end{figure}

Using the same hyperparameters, \Cref{fig:limit} shows the median performance of \method on AAV and GFP w.r.t the ratio $r$ in both settings. In the \textit{random} setting, our method outperforms the best data point in the entire \textit{harder3} dataset using only $20\%$ of the labeled data, which amounts to fewer than $100$ samples per set. In the \textit{lowest} setting, the method requires $50\%$ of the labeled data, approximately $200$ samples per set, to achieve similar performance. This demonstrates that our method is effective even under extreme conditions with limited labeled data.

\section{Conclusion} \label{sec:conclusion}
This paper presents a novel method, \methodname, to address the scarcity of labeled data in practical protein sequence design. \method embeds protein sequences into a continuous latent space. In this space, we generate additional synthetic latent points by interpolating existing data points and Gaussian noise and estimating their fitness scores using label propagation. We provide theoretical underpinnings based on convex hull and extrapolation to support \methodname's efficacy. Experimental results confirm the method's effectiveness in protein design under extreme labeled data-limited conditions. 
\paragraph{\textbf{Limitations}} Our proposed method, \methodname, has some limitations on hyperparameters. These numbers are sensitive to landscape characteristics, and one should search for their optimal values when optimizing new landscapes. We have detailed the hyperparameter tuning process in \Cref{app:method-hparams} and found that, despite some sensitivity differences between datasets, the optimal hyperparameters remain relatively stable across multiple difficulty levels within the same landscape, thereby reducing the burden of searching for optimal settings. Additionally, we theoretically characterize the relationship between \methodname's hyperparameters and the sampled nodes, guaranteeing controllability over the entire framework. A promising extension of \method is to find better ways to determine the optimal value of $\beta$, which can be done by analyzing the effect of random mutations on each family of protein sequences.


\bibliographystyle{ACM-Reference-Format}
\bibliography{main}


\begin{thebibliography}{45}


\ifx \showCODEN    \undefined \def \showCODEN     #1{\unskip}     \fi
\ifx \showDOI      \undefined \def \showDOI       #1{#1}\fi
\ifx \showISBNx    \undefined \def \showISBNx     #1{\unskip}     \fi
\ifx \showISBNxiii \undefined \def \showISBNxiii  #1{\unskip}     \fi
\ifx \showISSN     \undefined \def \showISSN      #1{\unskip}     \fi
\ifx \showLCCN     \undefined \def \showLCCN      #1{\unskip}     \fi
\ifx \shownote     \undefined \def \shownote      #1{#1}          \fi
\ifx \showarticletitle \undefined \def \showarticletitle #1{#1}   \fi
\ifx \showURL      \undefined \def \showURL       {\relax}        \fi
\providecommand\bibfield[2]{#2}
\providecommand\bibinfo[2]{#2}
\providecommand\natexlab[1]{#1}
\providecommand\showeprint[2][]{arXiv:#2}

\bibitem[Ahn et~al\mbox{.}(2020)]%
        {pmlr-v100-ahn20a}
\bibfield{author}{\bibinfo{person}{Michael Ahn}, \bibinfo{person}{Henry Zhu},
  \bibinfo{person}{Kristian Hartikainen}, \bibinfo{person}{Hugo Ponte},
  \bibinfo{person}{Abhishek Gupta}, \bibinfo{person}{Sergey Levine}, {and}
  \bibinfo{person}{Vikash Kumar}.} \bibinfo{year}{2020}\natexlab{}.
\newblock \showarticletitle{ROBEL: Robotics Benchmarks for Learning with
  Low-Cost Robots}. In \bibinfo{booktitle}{\emph{Proceedings of the Conference
  on Robot Learning}} \emph{(\bibinfo{series}{Proceedings of Machine Learning
  Research}, Vol.~\bibinfo{volume}{100})},
  \bibfield{editor}{\bibinfo{person}{Leslie~Pack Kaelbling},
  \bibinfo{person}{Danica Kragic}, {and} \bibinfo{person}{Komei Sugiura}}
  (Eds.). \bibinfo{publisher}{PMLR}, \bibinfo{pages}{1300--1313}.
\newblock
\urldef\tempurl%
\url{https://proceedings.mlr.press/v100/ahn20a.html}
\showURL{%
\tempurl}


\bibitem[Akiba et~al\mbox{.}(2019)]%
        {10.1145/3292500.3330701}
\bibfield{author}{\bibinfo{person}{Takuya Akiba}, \bibinfo{person}{Shotaro
  Sano}, \bibinfo{person}{Toshihiko Yanase}, \bibinfo{person}{Takeru Ohta},
  {and} \bibinfo{person}{Masanori Koyama}.} \bibinfo{year}{2019}\natexlab{}.
\newblock \showarticletitle{Optuna: A Next-generation Hyperparameter
  Optimization Framework}. In \bibinfo{booktitle}{\emph{Proceedings of the 25th
  ACM SIGKDD International Conference on Knowledge Discovery \& Data Mining}}
  (Anchorage, AK, USA) \emph{(\bibinfo{series}{KDD '19})}.
  \bibinfo{publisher}{Association for Computing Machinery},
  \bibinfo{address}{New York, NY, USA}, \bibinfo{pages}{2623–2631}.
\newblock
\showISBNx{9781450362016}
\urldef\tempurl%
\url{https://doi.org/10.1145/3292500.3330701}
\showDOI{\tempurl}


\bibitem[Arnold(1996)]%
        {ARNOLD19965091}
\bibfield{author}{\bibinfo{person}{Frances~H. Arnold}.}
  \bibinfo{year}{1996}\natexlab{}.
\newblock \showarticletitle{Directed evolution: Creating biocatalysts for the
  future}.
\newblock \bibinfo{journal}{\emph{Chemical Engineering Science}}
  \bibinfo{volume}{51}, \bibinfo{number}{23} (\bibinfo{year}{1996}),
  \bibinfo{pages}{5091--5102}.
\newblock
\showISSN{0009-2509}
\urldef\tempurl%
\url{https://doi.org/10.1016/S0009-2509(96)00288-6}
\showDOI{\tempurl}


\bibitem[Arnold(2018)]%
        {https://doi.org/10.1002/anie.201708408}
\bibfield{author}{\bibinfo{person}{Frances~H. Arnold}.}
  \bibinfo{year}{2018}\natexlab{}.
\newblock \showarticletitle{Directed Evolution: Bringing New Chemistry to
  Life}.
\newblock \bibinfo{journal}{\emph{Angewandte Chemie International Edition}}
  \bibinfo{volume}{57}, \bibinfo{number}{16} (\bibinfo{year}{2018}),
  \bibinfo{pages}{4143--4148}.
\newblock
\urldef\tempurl%
\url{https://doi.org/10.1002/anie.201708408}
\showDOI{\tempurl}
\showeprint{https://onlinelibrary.wiley.com/doi/pdf/10.1002/anie.201708408}


\bibitem[Balestriero et~al\mbox{.}(2021)]%
        {extrapolate}
\bibfield{author}{\bibinfo{person}{Randall Balestriero},
  \bibinfo{person}{Jerome Pesenti}, {and} \bibinfo{person}{Yann LeCun}.}
  \bibinfo{year}{2021}\natexlab{}.
\newblock \showarticletitle{Learning in high dimension always amounts to
  extrapolation}.
\newblock \bibinfo{journal}{\emph{arXiv preprint arXiv:2110.09485}}
  (\bibinfo{year}{2021}).
\newblock


\bibitem[Brockman et~al\mbox{.}(2016)]%
        {brockman2016openaigym}
\bibfield{author}{\bibinfo{person}{Greg Brockman}, \bibinfo{person}{Vicki
  Cheung}, \bibinfo{person}{Ludwig Pettersson}, \bibinfo{person}{Jonas
  Schneider}, \bibinfo{person}{John Schulman}, \bibinfo{person}{Jie Tang},
  {and} \bibinfo{person}{Wojciech Zaremba}.} \bibinfo{year}{2016}\natexlab{}.
\newblock \bibinfo{title}{OpenAI Gym}.
\newblock
\newblock
\showeprint[arxiv]{1606.01540}~[cs.LG]
\urldef\tempurl%
\url{https://arxiv.org/abs/1606.01540}
\showURL{%
\tempurl}


\bibitem[Brookes et~al\mbox{.}(2019)]%
        {brookes2019conditioning}
\bibfield{author}{\bibinfo{person}{David Brookes}, \bibinfo{person}{Hahnbeom
  Park}, {and} \bibinfo{person}{Jennifer Listgarten}.}
  \bibinfo{year}{2019}\natexlab{}.
\newblock \showarticletitle{Conditioning by adaptive sampling for robust
  design}. In \bibinfo{booktitle}{\emph{Proceedings of the 36th International
  Conference on Machine Learning}} \emph{(\bibinfo{series}{Proceedings of
  Machine Learning Research}, Vol.~\bibinfo{volume}{97})},
  \bibfield{editor}{\bibinfo{person}{Kamalika Chaudhuri} {and}
  \bibinfo{person}{Ruslan Salakhutdinov}} (Eds.). \bibinfo{publisher}{PMLR},
  \bibinfo{pages}{773--782}.
\newblock
\urldef\tempurl%
\url{https://proceedings.mlr.press/v97/brookes19a.html}
\showURL{%
\tempurl}


\bibitem[Brookes et~al\mbox{.}(2022a)]%
        {doi:10.1073/pnas.2109649118}
\bibfield{author}{\bibinfo{person}{David~H. Brookes}, \bibinfo{person}{Amirali
  Aghazadeh}, {and} \bibinfo{person}{Jennifer Listgarten}.}
  \bibinfo{year}{2022}\natexlab{a}.
\newblock \showarticletitle{On the sparsity of fitness functions and
  implications for learning}.
\newblock \bibinfo{journal}{\emph{Proceedings of the National Academy of
  Sciences}} \bibinfo{volume}{119}, \bibinfo{number}{1} (\bibinfo{year}{2022}),
  \bibinfo{pages}{e2109649118}.
\newblock
\urldef\tempurl%
\url{https://doi.org/10.1073/pnas.2109649118}
\showDOI{\tempurl}
\showeprint{https://www.pnas.org/doi/pdf/10.1073/pnas.2109649118}


\bibitem[Brookes et~al\mbox{.}(2022b)]%
        {brookes2022sparsity}
\bibfield{author}{\bibinfo{person}{David~H Brookes}, \bibinfo{person}{Amirali
  Aghazadeh}, {and} \bibinfo{person}{Jennifer Listgarten}.}
  \bibinfo{year}{2022}\natexlab{b}.
\newblock \showarticletitle{On the sparsity of fitness functions and
  implications for learning}.
\newblock \bibinfo{journal}{\emph{Proceedings of the National Academy of
  Sciences}} \bibinfo{volume}{119}, \bibinfo{number}{1} (\bibinfo{year}{2022}),
  \bibinfo{pages}{e2109649118}.
\newblock


\bibitem[Brown et~al\mbox{.}(2020)]%
        {gpt}
\bibfield{author}{\bibinfo{person}{Tom Brown}, \bibinfo{person}{Benjamin Mann},
  \bibinfo{person}{Nick Ryder}, \bibinfo{person}{Melanie Subbiah},
  \bibinfo{person}{Jared~D Kaplan}, \bibinfo{person}{Prafulla Dhariwal},
  \bibinfo{person}{Arvind Neelakantan}, \bibinfo{person}{Pranav Shyam},
  \bibinfo{person}{Girish Sastry}, \bibinfo{person}{Amanda Askell},
  \bibinfo{person}{Sandhini Agarwal}, \bibinfo{person}{Ariel Herbert-Voss},
  \bibinfo{person}{Gretchen Krueger}, \bibinfo{person}{Tom Henighan},
  \bibinfo{person}{Rewon Child}, \bibinfo{person}{Aditya Ramesh},
  \bibinfo{person}{Daniel Ziegler}, \bibinfo{person}{Jeffrey Wu},
  \bibinfo{person}{Clemens Winter}, \bibinfo{person}{Chris Hesse},
  \bibinfo{person}{Mark Chen}, \bibinfo{person}{Eric Sigler},
  \bibinfo{person}{Mateusz Litwin}, \bibinfo{person}{Scott Gray},
  \bibinfo{person}{Benjamin Chess}, \bibinfo{person}{Jack Clark},
  \bibinfo{person}{Christopher Berner}, \bibinfo{person}{Sam McCandlish},
  \bibinfo{person}{Alec Radford}, \bibinfo{person}{Ilya Sutskever}, {and}
  \bibinfo{person}{Dario Amodei}.} \bibinfo{year}{2020}\natexlab{}.
\newblock \showarticletitle{Language Models are Few-Shot Learners}. In
  \bibinfo{booktitle}{\emph{Advances in Neural Information Processing
  Systems}}, \bibfield{editor}{\bibinfo{person}{H.~Larochelle},
  \bibinfo{person}{M.~Ranzato}, \bibinfo{person}{R.~Hadsell},
  \bibinfo{person}{M.F. Balcan}, {and} \bibinfo{person}{H.~Lin}} (Eds.),
  Vol.~\bibinfo{volume}{33}. \bibinfo{publisher}{Curran Associates, Inc.},
  \bibinfo{pages}{1877--1901}.
\newblock
\urldef\tempurl%
\url{https://proceedings.neurips.cc/paper_files/paper/2020/file/1457c0d6bfcb4967418bfb8ac142f64a-Paper.pdf}
\showURL{%
\tempurl}


\bibitem[Bryant et~al\mbox{.}(2021)]%
        {bryant2021deep}
\bibfield{author}{\bibinfo{person}{Drew~H. Bryant}, \bibinfo{person}{Ali
  Bashir}, \bibinfo{person}{Sam Sinai}, \bibinfo{person}{Nina~K. Jain},
  \bibinfo{person}{Pierce~J. Ogden}, \bibinfo{person}{Patrick~F. Riley},
  \bibinfo{person}{George~M. Church}, \bibinfo{person}{Lucy~J. Colwell}, {and}
  \bibinfo{person}{Eric~D. Kelsic}.} \bibinfo{year}{2021}\natexlab{}.
\newblock \showarticletitle{Deep diversification of an AAV capsid protein by
  machine learning}.
\newblock \bibinfo{journal}{\emph{Nature Biotechnology}} \bibinfo{volume}{39},
  \bibinfo{number}{6} (\bibinfo{date}{01 Jun} \bibinfo{year}{2021}),
  \bibinfo{pages}{691--696}.
\newblock
\showISSN{1546-1696}
\urldef\tempurl%
\url{https://doi.org/10.1038/s41587-020-00793-4}
\showDOI{\tempurl}


\bibitem[Castro et~al\mbox{.}(2022)]%
        {Castro2022}
\bibfield{author}{\bibinfo{person}{Egbert Castro}, \bibinfo{person}{Abhinav
  Godavarthi}, \bibinfo{person}{Julian Rubinfien}, \bibinfo{person}{Kevin
  Givechian}, \bibinfo{person}{Dhananjay Bhaskar}, {and} \bibinfo{person}{Smita
  Krishnaswamy}.} \bibinfo{year}{2022}\natexlab{}.
\newblock \showarticletitle{Transformer-based protein generation with
  regularized latent space optimization}.
\newblock \bibinfo{journal}{\emph{Nature Machine Intelligence}}
  \bibinfo{volume}{4}, \bibinfo{number}{10} (\bibinfo{date}{01 Oct}
  \bibinfo{year}{2022}), \bibinfo{pages}{840--851}.
\newblock
\showISSN{2522-5839}
\urldef\tempurl%
\url{https://doi.org/10.1038/s42256-022-00532-1}
\showDOI{\tempurl}


\bibitem[Chandrasekaran et~al\mbox{.}(2012)]%
        {expectation_norm_bound}
\bibfield{author}{\bibinfo{person}{Venkat Chandrasekaran},
  \bibinfo{person}{Benjamin Recht}, \bibinfo{person}{Pablo~A Parrilo}, {and}
  \bibinfo{person}{Alan~S Willsky}.} \bibinfo{year}{2012}\natexlab{}.
\newblock \showarticletitle{The convex geometry of linear inverse problems}.
\newblock \bibinfo{journal}{\emph{Foundations of Computational mathematics}}
  \bibinfo{volume}{12}, \bibinfo{number}{6} (\bibinfo{year}{2012}),
  \bibinfo{pages}{805--849}.
\newblock


\bibitem[Chen et~al\mbox{.}(2022)]%
        {NEURIPS2022_bd391cf5}
\bibfield{author}{\bibinfo{person}{Can Chen}, \bibinfo{person}{Yingxueff
  Zhang}, \bibinfo{person}{Jie Fu}, \bibinfo{person}{Xue~(Steve) Liu}, {and}
  \bibinfo{person}{Mark Coates}.} \bibinfo{year}{2022}\natexlab{}.
\newblock \showarticletitle{Bidirectional Learning for Offline Infinite-width
  Model-based Optimization}. In \bibinfo{booktitle}{\emph{Advances in Neural
  Information Processing Systems}},
  \bibfield{editor}{\bibinfo{person}{S.~Koyejo}, \bibinfo{person}{S.~Mohamed},
  \bibinfo{person}{A.~Agarwal}, \bibinfo{person}{D.~Belgrave},
  \bibinfo{person}{K.~Cho}, {and} \bibinfo{person}{A.~Oh}} (Eds.),
  Vol.~\bibinfo{volume}{35}. \bibinfo{publisher}{Curran Associates, Inc.},
  \bibinfo{pages}{29454--29467}.
\newblock
\urldef\tempurl%
\url{https://proceedings.neurips.cc/paper_files/paper/2022/file/bd391cf5bdc4b63674d6da3edc1bde0d-Paper-Conference.pdf}
\showURL{%
\tempurl}


\bibitem[Chen et~al\mbox{.}(2023)]%
        {chen2023ampdiffusion}
\bibfield{author}{\bibinfo{person}{Tianlai Chen}, \bibinfo{person}{Pranay
  Vure}, \bibinfo{person}{Rishab Pulugurta}, {and} \bibinfo{person}{Pranam
  Chatterjee}.} \bibinfo{year}{2023}\natexlab{}.
\newblock \showarticletitle{{AMP}-Diffusion: Integrating Latent Diffusion with
  Protein Language Models for Antimicrobial Peptide Generation}. In
  \bibinfo{booktitle}{\emph{NeurIPS 2023 Generative AI and Biology (GenBio)
  Workshop}}.
\newblock
\urldef\tempurl%
\url{https://openreview.net/forum?id=145TM9VQhx}
\showURL{%
\tempurl}


\bibitem[Emami et~al\mbox{.}(2023)]%
        {Emami2023}
\bibfield{author}{\bibinfo{person}{Patrick Emami}, \bibinfo{person}{Aidan
  Perreault}, \bibinfo{person}{Jeffrey Law}, \bibinfo{person}{David Biagioni},
  {and} \bibinfo{person}{Peter St.~John}.} \bibinfo{year}{2023}\natexlab{}.
\newblock \showarticletitle{Plug and play directed evolution of proteins with
  gradient-based discrete MCMC}.
\newblock \bibinfo{journal}{\emph{Machine Learning: Science and Technology}}
  \bibinfo{volume}{4}, \bibinfo{number}{2} (\bibinfo{date}{April}
  \bibinfo{year}{2023}), \bibinfo{pages}{025014}.
\newblock
\showISSN{2632-2153}
\urldef\tempurl%
\url{https://doi.org/10.1088/2632-2153/accacd}
\showDOI{\tempurl}


\bibitem[Fowler and Fields(2014)]%
        {Fowler2014}
\bibfield{author}{\bibinfo{person}{Douglas~M. Fowler} {and}
  \bibinfo{person}{Stanley Fields}.} \bibinfo{year}{2014}\natexlab{}.
\newblock \showarticletitle{Deep mutational scanning: a new style of protein
  science}.
\newblock \bibinfo{journal}{\emph{Nature Methods}} \bibinfo{volume}{11},
  \bibinfo{number}{8} (\bibinfo{date}{01 Aug} \bibinfo{year}{2014}),
  \bibinfo{pages}{801--807}.
\newblock
\showISSN{1548-7105}
\urldef\tempurl%
\url{https://doi.org/10.1038/nmeth.3027}
\showDOI{\tempurl}


\bibitem[Frey et~al\mbox{.}(2024)]%
        {frey2024protein}
\bibfield{author}{\bibinfo{person}{Nathan~C. Frey}, \bibinfo{person}{Dan
  Berenberg}, \bibinfo{person}{Karina Zadorozhny}, \bibinfo{person}{Joseph
  Kleinhenz}, \bibinfo{person}{Julien Lafrance-Vanasse},
  \bibinfo{person}{Isidro Hotzel}, \bibinfo{person}{Yan Wu},
  \bibinfo{person}{Stephen Ra}, \bibinfo{person}{Richard Bonneau},
  \bibinfo{person}{Kyunghyun Cho}, \bibinfo{person}{Andreas Loukas},
  \bibinfo{person}{Vladimir Gligorijevic}, {and} \bibinfo{person}{Saeed
  Saremi}.} \bibinfo{year}{2024}\natexlab{}.
\newblock \showarticletitle{Protein Discovery with Discrete Walk-Jump
  Sampling}. In \bibinfo{booktitle}{\emph{The Twelfth International Conference
  on Learning Representations}}.
\newblock
\urldef\tempurl%
\url{https://openreview.net/forum?id=zMPHKOmQNb}
\showURL{%
\tempurl}


\bibitem[Gómez-Bombarelli et~al\mbox{.}(2018)]%
        {doi:10.1021/acscentsci.7b00572}
\bibfield{author}{\bibinfo{person}{Rafael Gómez-Bombarelli},
  \bibinfo{person}{Jennifer~N. Wei}, \bibinfo{person}{David Duvenaud},
  \bibinfo{person}{José~Miguel Hernández-Lobato}, \bibinfo{person}{Benjamín
  Sánchez-Lengeling}, \bibinfo{person}{Dennis Sheberla},
  \bibinfo{person}{Jorge Aguilera-Iparraguirre}, \bibinfo{person}{Timothy~D.
  Hirzel}, \bibinfo{person}{Ryan~P. Adams}, {and} \bibinfo{person}{Alán
  Aspuru-Guzik}.} \bibinfo{year}{2018}\natexlab{}.
\newblock \showarticletitle{Automatic Chemical Design Using a Data-Driven
  Continuous Representation of Molecules}.
\newblock \bibinfo{journal}{\emph{ACS Central Science}} \bibinfo{volume}{4},
  \bibinfo{number}{2} (\bibinfo{year}{2018}), \bibinfo{pages}{268--276}.
\newblock
\urldef\tempurl%
\url{https://doi.org/10.1021/acscentsci.7b00572}
\showDOI{\tempurl}
\showeprint{https://doi.org/10.1021/acscentsci.7b00572}
\newblock
\shownote{PMID: 29532027}.


\bibitem[Huang et~al\mbox{.}(2016)]%
        {Huang2016}
\bibfield{author}{\bibinfo{person}{Po-Ssu Huang}, \bibinfo{person}{Scott~E.
  Boyken}, {and} \bibinfo{person}{David Baker}.}
  \bibinfo{year}{2016}\natexlab{}.
\newblock \showarticletitle{The coming of age of de novo protein design}.
\newblock \bibinfo{journal}{\emph{Nature}} \bibinfo{volume}{537},
  \bibinfo{number}{7620} (\bibinfo{date}{01 Sep} \bibinfo{year}{2016}),
  \bibinfo{pages}{320--327}.
\newblock
\showISSN{1476-4687}
\urldef\tempurl%
\url{https://doi.org/10.1038/nature19946}
\showDOI{\tempurl}


\bibitem[Huang et~al\mbox{.}(2021)]%
        {huang2021combining}
\bibfield{author}{\bibinfo{person}{Qian Huang}, \bibinfo{person}{Horace He},
  \bibinfo{person}{Abhay Singh}, \bibinfo{person}{Ser-Nam Lim}, {and}
  \bibinfo{person}{Austin Benson}.} \bibinfo{year}{2021}\natexlab{}.
\newblock \showarticletitle{Combining Label Propagation and Simple Models
  out-performs Graph Neural Networks}. In
  \bibinfo{booktitle}{\emph{International Conference on Learning
  Representations}}.
\newblock
\urldef\tempurl%
\url{https://openreview.net/forum?id=8E1-f3VhX1o}
\showURL{%
\tempurl}


\bibitem[Jain et~al\mbox{.}(2022)]%
        {pmlr-v162-jain22a}
\bibfield{author}{\bibinfo{person}{Moksh Jain}, \bibinfo{person}{Emmanuel
  Bengio}, \bibinfo{person}{Alex Hernandez-Garcia}, \bibinfo{person}{Jarrid
  Rector-Brooks}, \bibinfo{person}{Bonaventure F.~P. Dossou},
  \bibinfo{person}{Chanakya~Ajit Ekbote}, \bibinfo{person}{Jie Fu},
  \bibinfo{person}{Tianyu Zhang}, \bibinfo{person}{Michael Kilgour},
  \bibinfo{person}{Dinghuai Zhang}, \bibinfo{person}{Lena Simine},
  \bibinfo{person}{Payel Das}, {and} \bibinfo{person}{Yoshua Bengio}.}
  \bibinfo{year}{2022}\natexlab{}.
\newblock \showarticletitle{Biological Sequence Design with {GF}low{N}ets}. In
  \bibinfo{booktitle}{\emph{Proceedings of the 39th International Conference on
  Machine Learning}} \emph{(\bibinfo{series}{Proceedings of Machine Learning
  Research}, Vol.~\bibinfo{volume}{162})},
  \bibfield{editor}{\bibinfo{person}{Kamalika Chaudhuri},
  \bibinfo{person}{Stefanie Jegelka}, \bibinfo{person}{Le~Song},
  \bibinfo{person}{Csaba Szepesvari}, \bibinfo{person}{Gang Niu}, {and}
  \bibinfo{person}{Sivan Sabato}} (Eds.). \bibinfo{publisher}{PMLR},
  \bibinfo{pages}{9786--9801}.
\newblock
\urldef\tempurl%
\url{https://proceedings.mlr.press/v162/jain22a.html}
\showURL{%
\tempurl}


\bibitem[Kauffman and Weinberger(1989)]%
        {Kauffman1989}
\bibfield{author}{\bibinfo{person}{Stuart~A. Kauffman} {and}
  \bibinfo{person}{Edward~D. Weinberger}.} \bibinfo{year}{1989}\natexlab{}.
\newblock \showarticletitle{The NK model of rugged fitness landscapes and its
  application to maturation of the immune response}.
\newblock \bibinfo{journal}{\emph{Journal of Theoretical Biology}}
  \bibinfo{volume}{141}, \bibinfo{number}{2} (\bibinfo{date}{Nov.}
  \bibinfo{year}{1989}), \bibinfo{pages}{211–245}.
\newblock
\showISSN{0022-5193}
\urldef\tempurl%
\url{https://doi.org/10.1016/s0022-5193(89)80019-0}
\showDOI{\tempurl}


\bibitem[Kirillov et~al\mbox{.}(2023)]%
        {sam}
\bibfield{author}{\bibinfo{person}{Alexander Kirillov}, \bibinfo{person}{Eric
  Mintun}, \bibinfo{person}{Nikhila Ravi}, \bibinfo{person}{Hanzi Mao},
  \bibinfo{person}{Chloe Rolland}, \bibinfo{person}{Laura Gustafson},
  \bibinfo{person}{Tete Xiao}, \bibinfo{person}{Spencer Whitehead},
  \bibinfo{person}{Alexander~C. Berg}, \bibinfo{person}{Wan-Yen Lo},
  \bibinfo{person}{Piotr Doll{\'a}r}, {and} \bibinfo{person}{Ross Girshick}.}
  \bibinfo{year}{2023}\natexlab{}.
\newblock \showarticletitle{Segment Anything}.
\newblock \bibinfo{journal}{\emph{arXiv:2304.02643}} (\bibinfo{year}{2023}).
\newblock


\bibitem[Kirjner et~al\mbox{.}(2024)]%
        {kirjner2024improving}
\bibfield{author}{\bibinfo{person}{Andrew Kirjner}, \bibinfo{person}{Jason
  Yim}, \bibinfo{person}{Raman Samusevich}, \bibinfo{person}{Shahar Bracha},
  \bibinfo{person}{Tommi~S. Jaakkola}, \bibinfo{person}{Regina Barzilay}, {and}
  \bibinfo{person}{Ila~R Fiete}.} \bibinfo{year}{2024}\natexlab{}.
\newblock \showarticletitle{Improving protein optimization with smoothed
  fitness landscapes}. In \bibinfo{booktitle}{\emph{The Twelfth International
  Conference on Learning Representations}}.
\newblock
\urldef\tempurl%
\url{https://openreview.net/forum?id=rxlF2Zv8x0}
\showURL{%
\tempurl}


\bibitem[Kumar and Levine(2020)]%
        {NEURIPS2020_373e4c5d}
\bibfield{author}{\bibinfo{person}{Aviral Kumar} {and} \bibinfo{person}{Sergey
  Levine}.} \bibinfo{year}{2020}\natexlab{}.
\newblock \showarticletitle{Model Inversion Networks for Model-Based
  Optimization}. In \bibinfo{booktitle}{\emph{Advances in Neural Information
  Processing Systems}}, \bibfield{editor}{\bibinfo{person}{H.~Larochelle},
  \bibinfo{person}{M.~Ranzato}, \bibinfo{person}{R.~Hadsell},
  \bibinfo{person}{M.F. Balcan}, {and} \bibinfo{person}{H.~Lin}} (Eds.),
  Vol.~\bibinfo{volume}{33}. \bibinfo{publisher}{Curran Associates, Inc.},
  \bibinfo{pages}{5126--5137}.
\newblock
\urldef\tempurl%
\url{https://proceedings.neurips.cc/paper_files/paper/2020/file/373e4c5d8edfa8b74fd4b6791d0cf6dc-Paper.pdf}
\showURL{%
\tempurl}


\bibitem[Lee et~al\mbox{.}(2023)]%
        {lee2023protein}
\bibfield{author}{\bibinfo{person}{Minji Lee}, \bibinfo{person}{Luiz~Felipe
  Vecchietti}, \bibinfo{person}{Hyunkyu Jung}, \bibinfo{person}{Hyunjoo Ro},
  \bibinfo{person}{Meeyoung Cha}, {and} \bibinfo{person}{Ho~Min Kim}.}
  \bibinfo{year}{2023}\natexlab{}.
\newblock \bibinfo{title}{Protein Sequence Design in a Latent Space via
  Model-based Reinforcement Learning}.
\newblock
\newblock
\urldef\tempurl%
\url{https://openreview.net/forum?id=OhjGzRE5N6o}
\showURL{%
\tempurl}


\bibitem[Lee et~al\mbox{.}(2024)]%
        {lee2024robust}
\bibfield{author}{\bibinfo{person}{Minji Lee}, \bibinfo{person}{Luiz~Felipe
  Vecchietti}, \bibinfo{person}{Hyunkyu Jung}, \bibinfo{person}{Hyun~Joo Ro},
  \bibinfo{person}{Meeyoung Cha}, {and} \bibinfo{person}{Ho~Min Kim}.}
  \bibinfo{year}{2024}\natexlab{}.
\newblock \showarticletitle{Robust Optimization in Protein Fitness Landscapes
  Using Reinforcement Learning in Latent Space}. In
  \bibinfo{booktitle}{\emph{Forty-first International Conference on Machine
  Learning}}.
\newblock
\urldef\tempurl%
\url{https://openreview.net/forum?id=0zbxwvJqwf}
\showURL{%
\tempurl}


\bibitem[Lin et~al\mbox{.}(2023)]%
        {doi:10.1126/science.ade2574}
\bibfield{author}{\bibinfo{person}{Zeming Lin}, \bibinfo{person}{Halil Akin},
  \bibinfo{person}{Roshan Rao}, \bibinfo{person}{Brian Hie},
  \bibinfo{person}{Zhongkai Zhu}, \bibinfo{person}{Wenting Lu},
  \bibinfo{person}{Nikita Smetanin}, \bibinfo{person}{Robert Verkuil},
  \bibinfo{person}{Ori Kabeli}, \bibinfo{person}{Yaniv Shmueli},
  \bibinfo{person}{Allan dos Santos~Costa}, \bibinfo{person}{Maryam
  Fazel-Zarandi}, \bibinfo{person}{Tom Sercu}, \bibinfo{person}{Salvatore
  Candido}, {and} \bibinfo{person}{Alexander Rives}.}
  \bibinfo{year}{2023}\natexlab{}.
\newblock \showarticletitle{Evolutionary-scale prediction of atomic-level
  protein structure with a language model}.
\newblock \bibinfo{journal}{\emph{Science}} \bibinfo{volume}{379},
  \bibinfo{number}{6637} (\bibinfo{year}{2023}), \bibinfo{pages}{1123--1130}.
\newblock
\urldef\tempurl%
\url{https://doi.org/10.1126/science.ade2574}
\showDOI{\tempurl}
\showeprint{https://www.science.org/doi/pdf/10.1126/science.ade2574}


\bibitem[Liu and Nocedal(1989)]%
        {Liu1989}
\bibfield{author}{\bibinfo{person}{Dong~C. Liu} {and} \bibinfo{person}{Jorge
  Nocedal}.} \bibinfo{year}{1989}\natexlab{}.
\newblock \showarticletitle{On the limited memory BFGS method for large scale
  optimization}.
\newblock \bibinfo{journal}{\emph{Mathematical Programming}}
  \bibinfo{volume}{45}, \bibinfo{number}{1} (\bibinfo{date}{01 Aug}
  \bibinfo{year}{1989}), \bibinfo{pages}{503--528}.
\newblock
\showISSN{1436-4646}
\urldef\tempurl%
\url{https://doi.org/10.1007/BF01589116}
\showDOI{\tempurl}


\bibitem[Mashkaria et~al\mbox{.}(2023)]%
        {pmlr-v202-mashkaria23a}
\bibfield{author}{\bibinfo{person}{Satvik~Mehul Mashkaria},
  \bibinfo{person}{Siddarth Krishnamoorthy}, {and} \bibinfo{person}{Aditya
  Grover}.} \bibinfo{year}{2023}\natexlab{}.
\newblock \showarticletitle{Generative Pretraining for Black-Box Optimization}.
  In \bibinfo{booktitle}{\emph{Proceedings of the 40th International Conference
  on Machine Learning}} \emph{(\bibinfo{series}{Proceedings of Machine Learning
  Research}, Vol.~\bibinfo{volume}{202})},
  \bibfield{editor}{\bibinfo{person}{Andreas Krause}, \bibinfo{person}{Emma
  Brunskill}, \bibinfo{person}{Kyunghyun Cho}, \bibinfo{person}{Barbara
  Engelhardt}, \bibinfo{person}{Sivan Sabato}, {and} \bibinfo{person}{Jonathan
  Scarlett}} (Eds.). \bibinfo{publisher}{PMLR}, \bibinfo{pages}{24173--24197}.
\newblock
\urldef\tempurl%
\url{https://proceedings.mlr.press/v202/mashkaria23a.html}
\showURL{%
\tempurl}


\bibitem[Maynard~Smith(1970)]%
        {MAYNARDSMITH1970}
\bibfield{author}{\bibinfo{person}{John Maynard~Smith}.}
  \bibinfo{year}{1970}\natexlab{}.
\newblock \showarticletitle{Natural Selection and the Concept of a Protein
  Space}.
\newblock \bibinfo{journal}{\emph{Nature}} \bibinfo{volume}{225},
  \bibinfo{number}{5232} (\bibinfo{date}{Feb.} \bibinfo{year}{1970}),
  \bibinfo{pages}{563–564}.
\newblock
\showISSN{1476-4687}
\urldef\tempurl%
\url{https://doi.org/10.1038/225563a0}
\showDOI{\tempurl}


\bibitem[Nguyen et~al\mbox{.}(2023)]%
        {NEURIPS2023_8fab4407}
\bibfield{author}{\bibinfo{person}{Tung Nguyen}, \bibinfo{person}{Sudhanshu
  Agrawal}, {and} \bibinfo{person}{Aditya Grover}.}
  \bibinfo{year}{2023}\natexlab{}.
\newblock \showarticletitle{ExPT: Synthetic Pretraining for Few-Shot
  Experimental Design}. In \bibinfo{booktitle}{\emph{Advances in Neural
  Information Processing Systems}}, \bibfield{editor}{\bibinfo{person}{A.~Oh},
  \bibinfo{person}{T.~Naumann}, \bibinfo{person}{A.~Globerson},
  \bibinfo{person}{K.~Saenko}, \bibinfo{person}{M.~Hardt}, {and}
  \bibinfo{person}{S.~Levine}} (Eds.), Vol.~\bibinfo{volume}{36}.
  \bibinfo{publisher}{Curran Associates, Inc.}, \bibinfo{pages}{45856--45869}.
\newblock
\urldef\tempurl%
\url{https://proceedings.neurips.cc/paper_files/paper/2023/file/8fab4407e1fe9006b39180525c0d323c-Paper-Conference.pdf}
\showURL{%
\tempurl}


\bibitem[Qiu and Wei(2022)]%
        {Qiu_2022}
\bibfield{author}{\bibinfo{person}{Yuchi Qiu} {and} \bibinfo{person}{Guo-Wei
  Wei}.} \bibinfo{year}{2022}\natexlab{}.
\newblock \showarticletitle{CLADE 2.0: Evolution-Driven Cluster
  Learning-Assisted Directed Evolution}.
\newblock \bibinfo{journal}{\emph{Journal of Chemical Information and
  Modeling}} \bibinfo{volume}{62}, \bibinfo{number}{19} (\bibinfo{date}{Sept.}
  \bibinfo{year}{2022}), \bibinfo{pages}{4629–4641}.
\newblock
\showISSN{1549-960X}
\urldef\tempurl%
\url{https://doi.org/10.1021/acs.jcim.2c01046}
\showDOI{\tempurl}


\bibitem[Raghavan et~al\mbox{.}(2007)]%
        {near_linear}
\bibfield{author}{\bibinfo{person}{Usha~Nandini Raghavan},
  \bibinfo{person}{R\'eka Albert}, {and} \bibinfo{person}{Soundar Kumara}.}
  \bibinfo{year}{2007}\natexlab{}.
\newblock \showarticletitle{Near linear time algorithm to detect community
  structures in large-scale networks}.
\newblock \bibinfo{journal}{\emph{Phys. Rev. E}}  \bibinfo{volume}{76}
  (\bibinfo{date}{Sep} \bibinfo{year}{2007}), \bibinfo{pages}{036106}.
\newblock
Issue 3.
\urldef\tempurl%
\url{https://doi.org/10.1103/PhysRevE.76.036106}
\showDOI{\tempurl}


\bibitem[Ren et~al\mbox{.}(2022)]%
        {pmlr-v162-ren22a}
\bibfield{author}{\bibinfo{person}{Zhizhou Ren}, \bibinfo{person}{Jiahan Li},
  \bibinfo{person}{Fan Ding}, \bibinfo{person}{Yuan Zhou},
  \bibinfo{person}{Jianzhu Ma}, {and} \bibinfo{person}{Jian Peng}.}
  \bibinfo{year}{2022}\natexlab{}.
\newblock \showarticletitle{Proximal Exploration for Model-guided Protein
  Sequence Design}. In \bibinfo{booktitle}{\emph{Proceedings of the 39th
  International Conference on Machine Learning}}
  \emph{(\bibinfo{series}{Proceedings of Machine Learning Research},
  Vol.~\bibinfo{volume}{162})}, \bibfield{editor}{\bibinfo{person}{Kamalika
  Chaudhuri}, \bibinfo{person}{Stefanie Jegelka}, \bibinfo{person}{Le~Song},
  \bibinfo{person}{Csaba Szepesvari}, \bibinfo{person}{Gang Niu}, {and}
  \bibinfo{person}{Sivan Sabato}} (Eds.). \bibinfo{publisher}{PMLR},
  \bibinfo{pages}{18520--18536}.
\newblock
\urldef\tempurl%
\url{https://proceedings.mlr.press/v162/ren22a.html}
\showURL{%
\tempurl}


\bibitem[Sarkisyan et~al\mbox{.}(2016)]%
        {sarkisyan2016local}
\bibfield{author}{\bibinfo{person}{Karen~S. Sarkisyan},
  \bibinfo{person}{Dmitry~A. Bolotin}, \bibinfo{person}{Margarita~V. Meer},
  \bibinfo{person}{Dinara~R. Usmanova}, \bibinfo{person}{Alexander~S. Mishin},
  \bibinfo{person}{George~V. Sharonov}, \bibinfo{person}{Dmitry~N. Ivankov},
  \bibinfo{person}{Nina~G. Bozhanova}, \bibinfo{person}{Mikhail~S. Baranov},
  \bibinfo{person}{Onuralp Soylemez}, \bibinfo{person}{Natalya~S. Bogatyreva},
  \bibinfo{person}{Peter~K. Vlasov}, \bibinfo{person}{Evgeny~S. Egorov},
  \bibinfo{person}{Maria~D. Logacheva}, \bibinfo{person}{Alexey~S. Kondrashov},
  \bibinfo{person}{Dmitry~M. Chudakov}, \bibinfo{person}{Ekaterina~V.
  Putintseva}, \bibinfo{person}{Ilgar~Z. Mamedov}, \bibinfo{person}{Dan~S.
  Tawfik}, \bibinfo{person}{Konstantin~A. Lukyanov}, {and}
  \bibinfo{person}{Fyodor~A. Kondrashov}.} \bibinfo{year}{2016}\natexlab{}.
\newblock \showarticletitle{Local fitness landscape of the green fluorescent
  protein}.
\newblock \bibinfo{journal}{\emph{Nature}} \bibinfo{volume}{533},
  \bibinfo{number}{7603} (\bibinfo{date}{01 May} \bibinfo{year}{2016}),
  \bibinfo{pages}{397--401}.
\newblock
\showISSN{1476-4687}
\urldef\tempurl%
\url{https://doi.org/10.1038/nature17995}
\showDOI{\tempurl}


\bibitem[Sinai et~al\mbox{.}(2020)]%
        {DBLP:journals/corr/abs-2010-02141}
\bibfield{author}{\bibinfo{person}{Sam Sinai}, \bibinfo{person}{Richard Wang},
  \bibinfo{person}{Alexander Whatley}, \bibinfo{person}{Stewart Slocum},
  \bibinfo{person}{Elina Locane}, {and} \bibinfo{person}{Eric~D. Kelsic}.}
  \bibinfo{year}{2020}\natexlab{}.
\newblock \showarticletitle{AdaLead: {A} simple and robust adaptive greedy
  search algorithm for sequence design}.
\newblock \bibinfo{journal}{\emph{CoRR}}  \bibinfo{volume}{abs/2010.02141}
  (\bibinfo{year}{2020}).
\newblock
\showeprint[arXiv]{2010.02141}
\urldef\tempurl%
\url{https://arxiv.org/abs/2010.02141}
\showURL{%
\tempurl}


\bibitem[Stanton et~al\mbox{.}(2022)]%
        {pmlr-v162-stanton22a}
\bibfield{author}{\bibinfo{person}{Samuel Stanton}, \bibinfo{person}{Wesley
  Maddox}, \bibinfo{person}{Nate Gruver}, \bibinfo{person}{Phillip Maffettone},
  \bibinfo{person}{Emily Delaney}, \bibinfo{person}{Peyton Greenside}, {and}
  \bibinfo{person}{Andrew~Gordon Wilson}.} \bibinfo{year}{2022}\natexlab{}.
\newblock \showarticletitle{Accelerating {B}ayesian Optimization for Biological
  Sequence Design with Denoising Autoencoders}. In
  \bibinfo{booktitle}{\emph{Proceedings of the 39th International Conference on
  Machine Learning}} \emph{(\bibinfo{series}{Proceedings of Machine Learning
  Research}, Vol.~\bibinfo{volume}{162})},
  \bibfield{editor}{\bibinfo{person}{Kamalika Chaudhuri},
  \bibinfo{person}{Stefanie Jegelka}, \bibinfo{person}{Le~Song},
  \bibinfo{person}{Csaba Szepesvari}, \bibinfo{person}{Gang Niu}, {and}
  \bibinfo{person}{Sivan Sabato}} (Eds.). \bibinfo{publisher}{PMLR},
  \bibinfo{pages}{20459--20478}.
\newblock
\urldef\tempurl%
\url{https://proceedings.mlr.press/v162/stanton22a.html}
\showURL{%
\tempurl}


\bibitem[Trabucco et~al\mbox{.}(2022)]%
        {pmlr-v162-trabucco22a}
\bibfield{author}{\bibinfo{person}{Brandon Trabucco}, \bibinfo{person}{Xinyang
  Geng}, \bibinfo{person}{Aviral Kumar}, {and} \bibinfo{person}{Sergey
  Levine}.} \bibinfo{year}{2022}\natexlab{}.
\newblock \showarticletitle{Design-Bench: Benchmarks for Data-Driven Offline
  Model-Based Optimization}. In \bibinfo{booktitle}{\emph{Proceedings of the
  39th International Conference on Machine Learning}}
  \emph{(\bibinfo{series}{Proceedings of Machine Learning Research},
  Vol.~\bibinfo{volume}{162})}, \bibfield{editor}{\bibinfo{person}{Kamalika
  Chaudhuri}, \bibinfo{person}{Stefanie Jegelka}, \bibinfo{person}{Le~Song},
  \bibinfo{person}{Csaba Szepesvari}, \bibinfo{person}{Gang Niu}, {and}
  \bibinfo{person}{Sivan Sabato}} (Eds.). \bibinfo{publisher}{PMLR},
  \bibinfo{pages}{21658--21676}.
\newblock
\urldef\tempurl%
\url{https://proceedings.mlr.press/v162/trabucco22a.html}
\showURL{%
\tempurl}


\bibitem[Trabucco et~al\mbox{.}(2021)]%
        {pmlr-v139-trabucco21a}
\bibfield{author}{\bibinfo{person}{Brandon Trabucco}, \bibinfo{person}{Aviral
  Kumar}, \bibinfo{person}{Xinyang Geng}, {and} \bibinfo{person}{Sergey
  Levine}.} \bibinfo{year}{2021}\natexlab{}.
\newblock \showarticletitle{Conservative Objective Models for Effective Offline
  Model-Based Optimization}. In \bibinfo{booktitle}{\emph{Proceedings of the
  38th International Conference on Machine Learning}}
  \emph{(\bibinfo{series}{Proceedings of Machine Learning Research},
  Vol.~\bibinfo{volume}{139})}, \bibfield{editor}{\bibinfo{person}{Marina
  Meila} {and} \bibinfo{person}{Tong Zhang}} (Eds.). \bibinfo{publisher}{PMLR},
  \bibinfo{pages}{10358--10368}.
\newblock
\urldef\tempurl%
\url{https://proceedings.mlr.press/v139/trabucco21a.html}
\showURL{%
\tempurl}


\bibitem[Tran and Hy(2024)]%
        {Tran2023.11.28.568945}
\bibfield{author}{\bibinfo{person}{Thanh V.~T. Tran} {and}
  \bibinfo{person}{Truong~Son Hy}.} \bibinfo{year}{2024}\natexlab{}.
\newblock \showarticletitle{Protein Design by Directed Evolution Guided by
  Large Language Models}.
\newblock \bibinfo{journal}{\emph{bioRxiv}} (\bibinfo{year}{2024}).
\newblock
\urldef\tempurl%
\url{https://doi.org/10.1101/2023.11.28.568945}
\showDOI{\tempurl}
\showeprint{https://www.biorxiv.org/content/early/2024/05/02/2023.11.28.568945.full.pdf}


\bibitem[Wang et~al\mbox{.}(2021)]%
        {doi:10.1021/acs.chemrev.1c00260}
\bibfield{author}{\bibinfo{person}{Yajie Wang}, \bibinfo{person}{Pu Xue},
  \bibinfo{person}{Mingfeng Cao}, \bibinfo{person}{Tianhao Yu},
  \bibinfo{person}{Stephan~T. Lane}, {and} \bibinfo{person}{Huimin Zhao}.}
  \bibinfo{year}{2021}\natexlab{}.
\newblock \showarticletitle{Directed Evolution: Methodologies and
  Applications}.
\newblock \bibinfo{journal}{\emph{Chemical Reviews}} \bibinfo{volume}{121},
  \bibinfo{number}{20} (\bibinfo{year}{2021}), \bibinfo{pages}{12384--12444}.
\newblock
\urldef\tempurl%
\url{https://doi.org/10.1021/acs.chemrev.1c00260}
\showDOI{\tempurl}
\showeprint{https://doi.org/10.1021/acs.chemrev.1c00260}
\newblock
\shownote{PMID: 34297541}.


\bibitem[Wilson et~al\mbox{.}(2017)]%
        {wilson2017reparameterization}
\bibfield{author}{\bibinfo{person}{James~T. Wilson}, \bibinfo{person}{Riccardo
  Moriconi}, \bibinfo{person}{Frank Hutter}, {and} \bibinfo{person}{Marc~Peter
  Deisenroth}.} \bibinfo{year}{2017}\natexlab{}.
\newblock \bibinfo{title}{The reparameterization trick for acquisition
  functions}.
\newblock
\newblock
\showeprint[arxiv]{1712.00424}~[stat.ML]
\urldef\tempurl%
\url{https://arxiv.org/abs/1712.00424}
\showURL{%
\tempurl}


\bibitem[Zhou et~al\mbox{.}(2003)]%
        {NIPS2003_87682805}
\bibfield{author}{\bibinfo{person}{Dengyong Zhou}, \bibinfo{person}{Olivier
  Bousquet}, \bibinfo{person}{Thomas Lal}, \bibinfo{person}{Jason Weston},
  {and} \bibinfo{person}{Bernhard Sch\"{o}lkopf}.}
  \bibinfo{year}{2003}\natexlab{}.
\newblock \showarticletitle{Learning with Local and Global Consistency}. In
  \bibinfo{booktitle}{\emph{Advances in Neural Information Processing
  Systems}}, \bibfield{editor}{\bibinfo{person}{S.~Thrun},
  \bibinfo{person}{L.~Saul}, {and} \bibinfo{person}{B.~Sch\"{o}lkopf}} (Eds.),
  Vol.~\bibinfo{volume}{16}. \bibinfo{publisher}{MIT Press}.
\newblock
\urldef\tempurl%
\url{https://proceedings.neurips.cc/paper_files/paper/2003/file/87682805257e619d49b8e0dfdc14affa-Paper.pdf}
\showURL{%
\tempurl}


\end{thebibliography}

\appendix

\section{Proofs}\label{app:proof}

\subsection{Proof of \Cref{prop:prob_limit}}
\label{app:proof_of_prob_limit}
\begin{proof}
    We compute the probability that $z$ belongs to $\conv$, $\mathbf{P}(z \in \conv)$. By definition, we have
    \begin{align*}
        z &= \sum_{i=1}^n\lambda_ix_i\\
        \beta\overline{x} + (1-\beta)\epsilon =& \sum_{i=1}^n\lambda_ix_i\\
        \epsilon =& -\frac{\beta}{1 - \beta}\overline{x} + \sum_{i=1}^n\frac{\lambda_i}{1-\beta}x_i
    \end{align*}

Since $\overline{x}\in\mathbb{X}_d$, without loss of generality, let $\overline{x} = x_1$. The other cases can be computed similar, taking the mean gives the same results.
\begin{equation*}
    \epsilon = \frac{\lambda_1-\beta}{1-\beta}x_1 + \sum_{i=2}^n\frac{\lambda_i}{1-\beta} x_i 
\end{equation*}
This will lead to
\begin{align*}
    \|\epsilon\|^2 &= \frac{\lambda_1-\beta}{1-\beta}\langle \epsilon, x_1\rangle + \sum_{i=2}^n\frac{\lambda_i}{1-\beta}\langle \epsilon, x_i\rangle\\
    &\leq\underbrace{\frac{1+\beta}{1-\beta}}_{C_\beta}\max_{i=\overline{1,n}}|\langle \epsilon, x_i\rangle|.
\end{align*}

Since all $x_i$ are independent, for $x\in\mathcal{N}(0,I_d)$, we have:
$$\mathbf{P}\left(\|\epsilon\|^2 \leq C_{\beta}\max_{i=\overline{1,n}}|\langle \epsilon, x_i\rangle|\right) = 1- \lrbrac{\mathbf{P}(\|\epsilon\|^2 \geq C_{\beta}|\langle \epsilon, x   \rangle|)}^n.$$

Note that $\lrang{\epsilon,x} = \frac{1}{2}\left(\left\|\frac{\epsilon+x}{\sqrt{2}}\right\|^2 - \left\|\frac{\epsilon-x}{\sqrt{2}}\right\|^2 \right) = \frac{1}{2}(A-B)$ where $A,B\sim\chi_d^2$. From here, since $d$ is large, we consider Chi-square as the normal approximations
$$A,B,\|\epsilon\|^2\sim\mathcal{N}(d,2d).$$
We consider the two distribution 
\begin{align*}
    &S_1 = \|\epsilon\|^2 - \frac{C_{\beta}}{2}A + \frac{C_{\beta}}{2}B,\\
    &S_2 = \|\epsilon\|^2 + \frac{C_{\beta}}{2}A - \frac{C_{\beta}}{2}B.
\end{align*}
Since $A,B,\|\epsilon\|^2$ are all normal, $S_1$ and $S_2$ are also normal. Moreover,
$$\mathbb{E}S_1=\mathbb{E}S_2=d,$$
and the variances are
\begin{align*}
    \var(S_1) = \;&\frac{C_{\beta}^2}{4}\var(A) + \frac{C_{\beta}^2}{4}\var(B) +\var(\|\epsilon\|^2) - \frac{C_{\beta}^2}{2}\cov\lrbrac{A,B} \\
    &- C_{\beta}\cov\lrbrac{\|\epsilon\|^2, A} + C_{\beta}\cov\lrbrac{\|\epsilon\|^2, B}\\
    =\;&d(C_{\beta}^2+2)- \frac{C_{\beta}^2}{2}\cov\lrbrac{A,B}\\
    &- C_{\beta}\cov\lrbrac{\|\epsilon\|^2, A} + C_{\beta}\cov\lrbrac{\|\epsilon\|^2, B}\\
    \var(S_2) = \;&d(C_{\beta}^2+2)- \frac{C_{\beta}^2}{2}\cov\lrbrac{A,B}\\ 
    &+ C_{\beta}\cov\lrbrac{\|\epsilon\|^2, A} - C_{\beta}\cov\lrbrac{\|\epsilon\|^2, B}
\end{align*}

Due to symmetric distribution of $x$, we can see that $\cov\lrbrac{\|\epsilon\|^2, A} = \cov\lrbrac{\|\epsilon\|^2, B}$. Moreover, $\epsilon$ and $x$ are independent random vectors, thus $\epsilon + x$ and $\epsilon - x$ are also independent, thus $\cov(A,B)=0$. From these deductions, we have shown that

$$S_1,S_2\sim\mathcal{N}\lrbrac{d,d(C_{\beta}^2+2)}.$$

Since $\|\epsilon\|^2 - \frac{C_{\beta}}{2}|A-B|\geq 0$ iff $S_1,S_2\geq 0$, we derive that

\begin{align*}
    \mathbf{P}(\|\epsilon\|^2 \geq C_{\beta}|\langle \epsilon, x\rangle|)\;& = \mathbf{P}(S_1>0)\times\mathbf{P}(S_2>0)\\
    \;&=\Phi\lrbrac{\sqrt{\frac{d}{C^2_{\beta}+2}}}^2\\
    \;&\approx \lrbrac{1 - \frac{\exp\lrbrac{-d/(2(C^2_{\beta}+2))}}{\sqrt{2\pi d}(C^2_{\beta}+2)^{-1}}}^2
\end{align*}

Finally, we get that

\begin{align*}
    &\mathbf{P}\left(\|\epsilon\|^2 \leq C_{\beta}\max_{i=\overline{1,N}}|\langle \epsilon, x\rangle|\right)\\
    \approx\;& 1 - \lrbrac{1 - \frac{\exp\lrbrac{-\frac{d}{2(C^2_{\beta}+2)}}}{\sqrt{2\pi d}(C^2_{\beta}+2)^{-1}}}^{2N}\\
    =\;& 1 - \exp\lrbrac{2N\log\lrbrac{1 - \frac{\exp\lrbrac{-\frac{d}{2(C^2_{\beta}+2)}}}{\sqrt{2\pi d}(C^2_{\beta}+2)^{-1}}}}\\
    \approx\;& 1 -\exp\lrbrac{-2N\frac{\exp\lrbrac{-\frac{d}{2(C^2_{\beta}+2)}}}{\sqrt{2\pi d}(C^2_{\beta}+2)^{-1}}}
\end{align*}

Thus for all $N\ll \exp\lrbrac{\frac{d}{2(C^2_{\beta}+2)}}$ and $d$ sufficiently large, we have $$\mathbf{P}\left(\|\epsilon\|^2 \leq C_{\beta}\max_{i=\overline{1,N}}|\langle \epsilon, x\rangle|\right)\approx0$$

In other words,

$$\lim_{d\to\infty}\mathbf{P}(z\in \operatorname{Conv}(\mathbb{X}_d)) = 0.$$
\end{proof}

\subsection{Proof of \Cref{prop:distance}} 
\label{app:prop_distance}
\begin{proof} 
Since $D(z, \conv) = \underset{x \in \conv}{\inf} \big(\norm{z - x} \big)$, for any $\overline x \in \conv$, we should have: 
\begin{align*}
     \mathbb{E}[D(z, \conv)] \le \mathbb{E}\norm{z - \overline x}.
\end{align*}
Replace $z$ by $\beta * \overline x + (1-\beta) * \epsilon$, we obtain:
\begin{align*}
    \mathbb{E}\norm{z - \overline x} & \le \mathbb{E}\norm{\beta *\overline x + (1-\beta) * \epsilon - (1-\beta) * \overline x - \beta * \overline x} \\
    & \le \mathbb{E}\norm{(1-\beta) (\epsilon - \overline x)} \\
    & \le (1-\beta) (\mathbb{E}\norm{\epsilon} + \mathbb{E}\norm{\overline x})
\end{align*}
We have that $\epsilon \sim \mathcal{N}(0, I_d)$ and $\overline x \sim \mathcal{N}(0, I_d)$. \citet{expectation_norm_bound} has shown that the upper bound of their expectations' norms is $\sqrt{d}$. Thus, we have 
\begin{align*}
    \mathbb{E}\norm{z - \overline x} &\le (1-\beta) \big (\mathbb E \norm{\epsilon} + \mathbb E \norm{\overline x} \big)  \\
    & < (1-\beta) \big(\sqrt{d} + \sqrt{d}\big) = 2(1-\beta)\sqrt{d}.
\end{align*}
Therefore, 
\begin{equation*}
    \mathbb E [ D(z, \conv)] <  2(1-\beta)\sqrt{d},
\end{equation*}
which completes the proof.
\end{proof}

\section{Evaluation Metrics}\label{app:metrics}

We provide mathematical definitions for each metric. Note that \( \mathcal O_\psi \) is the evaluator provided by \citet{kirjner2024improving} to predict approximate fitness, serving as a proxy for experimental validation.

\begin{itemize}
    \item \textbf{(Normalized) Fitness} = $\text{median} \left( \left\{ \xi(\hat{s}_i; Y^*) \right\}_{i=1}^{N_{\text{samples}}} \right)$ where $\xi(\hat{s}; Y^*) = \frac{\mathcal O_\psi(\hat{s}) - \min(Y^*)}{\max(Y^*) - \min(Y^*)}$ is the min-max normalized fitness based on the lowest and highest known fitness in \( Y^* \).
    \item \textbf{Diversity} = $\text{median} \left( \left\{ \text{dist}(s, \hat{s}) : s, \hat{s} \in \hat{S}, s \neq \hat{s} \right\} \right)$ is the average sample similarity.
    \item \textbf{Novelty} = $\text{median} \left( \left\{ \eta(\hat{s}_i; S) \right\}_{i=1}^{N_{\text{samples}}} \right)$ where \\$\eta(s; S) = \min \left( \left\{ \text{dist}(s, \hat{s}) : \hat{s} \in S^*, \hat{s} \neq s \right\} \right)$ is the minimum distance of sample $s$ to any of the starting sequences $S$.
\end{itemize}

\begin{algorithm}[ht]
    \caption{\texttt{kNN}: k-Nearest Neighbors}
    \noindent 
    \begin{algorithmic}[1]
    \Require Current node $x$, All nodes $\mathcal V$
    \State $D(x) \leftarrow \bigcup_{x' \in \mathcal V/\{x\}} ||x' - x||$ \Comment{Euclidean distance.}
    \State $\mathcal X' \leftarrow \texttt{TopK}(D(x),\mathcal V)$ \Comment{Retrieve K closest nodes to $x$.}
    \State $\mathcal E(x) \leftarrow \bigcup_{x' \in \mathcal X'}(x,x')$ \Comment{Construct neighborhood around $x$.}
    \State \Return $\mathcal E(x)$
    \end{algorithmic}
    \label{algo:kNN}
\end{algorithm}

\section{Numerical Results For Other Protein Tasks} \label{app:protein}

In this section, we present experimental results of other difficulties of two protein datasets: GFP and AAV. \Cref{tab:app-stats} provides statistics for each level benchmark. We use the same settings as described in \Cref{sec:exp-setup} for this evaluation. \Cref{tab:app-numeric} outlined the results of baselines and our \methodname. Although our method is designed for scenarios with limited labeled data, it achieves state-of-the-art (SOTA) results across all difficulty levels in both datasets. Additionally, the enhanced version of ReLSO incorporating our smoothing strategy, S-ReLSO, outperforms the original version by a significant margin in all tasks, particularly in the GFP \textit{medium} task, where an 8-fold improvement is observed. These results demonstrate the effectiveness of our framework.

\begin{table}[ht]
\centering
\caption{Statistic of benchmarks.}{
    \label{tab:app-stats}
    \begin{tabular}{lccccc}
        \toprule
        \multirow{2}{*}{Task} & \multirow{2}{*}{Difficulty} & Fitness & Mutational & Best & \multirow{2}{*}{$|\mathcal D|$} \\
        & & Range ($\%$) & Gap & Fitness & \\
        \midrule
        \multirow{3}{*}{AAV} & Easy & $50-60$th & $0$ & $0.53$ & $5609$ \\
        & Medium & $20-40$th & $6$ & $0.38$ & $2828$ \\ 
        & Hard & $< 30$th & $7$ & $0.33$ & $2426$ \\
        \midrule
        \multirow{3}{*}{GFP} & Easy & $50-60$th & $0$ & $0.79$ & $4413$ \\
        & Medium & $20-40$th & $6$ & $0.62$ & $2139$ \\ 
        & Hard & $< 30$th & $7$ & $0.10$ & $3448$ \\
        \bottomrule
    \end{tabular}
}
\end{table}

\begin{table*}[ht]
    \centering
    \caption{\textbf{AAV and GFP optimization results} for \method and baseline methods. The standard deviation of 5 runs with different random seeds is indicated in parentheses.}
    \label{tab:app-numeric}
    \begin{tabular}{lccc|ccc|ccc}
    \toprule
         & \multicolumn{3}{c}{AAV \textit{easy} task} & \multicolumn{3}{c}{AAV \textit{medium} task} & \multicolumn{3}{c}{AAV \textit{hard} task} \\ 
         \cmidrule(lr){2-4} \cmidrule(lr){5-7} \cmidrule(lr){8-10}
         Method & Fitness $\uparrow$ & Diversity & Novelty & Fitness $\uparrow$ & Diversity & Novelty & Fitness $\uparrow$ & Diversity & Novelty \\
    \midrule
        AdaLead & 0.53 (0.0) & 5.7 (0.4) & 3.0 (0.7) & 0.49 (0.0) & 5.3 (0.7) & 6.3 (0.4) & 0.46 (0.0) & 5.4 (1.7) & 6.9 (0.9) \\
        CbAS & 0.03 (0.0) & 23.2 (0.2) & 17.5 (0.5) & 0.02 (0.0) & 23.1 (0.1) & 18.3 (0.4) & 0.02 (0.0) & 23.0 (0.2) & 18.5 (0.5) \\
        BO & 0.01 (0.0) & 20.1 (0.5) & 22.3 (0.4) & 0.00 (0.0) & 20.4 (0.2) & 21.5 (0.5) & 0.01 (0.0) & 20.5 (0.2) & 20.8 (0.4) \\
        GFN-AL & 0.05 (0.0) & 18.2 (3.3) & 21.0 (0.0) & 0.01 (0.0) & 16.9 (2.4) & 21.4 (1.0) & 0.01 (0.0) & 14.4 (6.2) & 21.6 (0.5) \\
        PEX & 0.44 (0.0) & 6.33 (1.1) & 4.0 (0.6) & 0.35 (0.0) & 6.86 (0.9) & 4.2 (1.0) & 0.29 (0.0) & 6.43 (0.5) & 3.8 (0.7) \\
        GGS & 0.49 (0.0) & 9.0 (0.2) & 8.0 (0.0) & 0.51 (0.0) & 4.9 (0.2) & 5.4 (0.5) & 0.60 (0.0) & 4.5 (0.5) & 7.0 (0.0) \\
        ReLSO & 0.18 (0.0) & 1.4 (0.0) & 5.0 (0.0) & 0.18 (0.0) & 15.7 (0.0) & 11.0 (0.0) & 0.03 (0.0) & 18.6 (0.0) & 15.0 (0.0) \\
        S-ReLSO & 0.47 (0.0) & 8.6 (0.0) & 4.0 (0.0) & 0.26 (0.0) & 13.7 (0.0) & 7.0 (0.0) & 0.2 (0.0) & 8.3 (0.0) & 11.0 (0.0) \\
        \textbf{\method} & \textbf{0.65 (0.0)} & 2.0 (0.7) & 1.0 (0.0) & \textbf{0.59 (0.0)} & 5.1 (2.3) & 5.4 (0.6) & \textbf{0.61 (0.0)} & 5.0 (1.0) & 7.0 (0.0) \\
    \toprule
        & \multicolumn{3}{c}{GFP \textit{easy} task} & \multicolumn{3}{c}{GFP \textit{medium} task} & \multicolumn{3}{c}{GFP \textit{hard} task} \\ 
         \cmidrule(lr){2-4} \cmidrule(lr){5-7} \cmidrule(lr){8-10}
         Method & Fitness $\uparrow$ & Diversity & Novelty & Fitness $\uparrow$ & Diversity & Novelty & Fitness $\uparrow$ & Diversity & Novelty \\
    \midrule
        AdaLead & 0.68 (0.0) & 7.0 (0.4) & 2.3 (0.4) & 0.52 (0.0) & 8.6 (3.0) & 4.5 (0.5) & 0.47 (0.0) & 8.3 (1.9) & 9.5 (2.7) \\
        CbAS & -0.09 (0.0) & 190.6 (1.5) & 171.5 (4.6) & -0.08 (0.0) & 171.5 (59.3) & 200.8 (2.2) & -0.09 (0.0) & 170.0 (29.2) & 202.3 (0.8) \\
        BO & -0.06 (0.0) & 55.7 (1.3) & 198.1 (5.6) & -0.04 (0.0) & 58.1 (2.0) & 199.8 (4.4) & -0.03 (0.1) & 59.4 (2.0) & 197.0 (7.6) \\
        GFN-AL & 0.27 (0.1) & 83.7 (83.2) & 222.7 (1.2) & 0.27 (0.1) & 65.5 (12.5) & 223.0 (1.0) & 0.19 (0.1) & 54.4 (32.7) & 224.0 (5.4) \\
        PEX & 0.62 (0.0) & 6.6 (0.3) & 3.9 (0.2) & 0.46 (0.0) & 7.8 (0.7) & 4.6 (1.0) & 0.13 (0.0) & 12.6 (1.2) & 7.1 (1.1) \\
        GGS & 0.84 (0.0) & 5.6 (0.2) & 3.5 (0.2) & 0.76 (0.0) & 3.7 (0.2) & 5.0 (0.0) & 0.74 (0.0) & 3.0 (0.1) & 8.0 (0.0) \\
        ReLSO & 0.40 (0.0) & 3.6 (0.0) & 6.0 (0.0) & 0.08 (0.0) & 29.0 (0.0) & 18.0 (0.0) & 0.60 (0.0) & 38.8 (0.0) & 8.0 (0.0) \\
        S-ReLSO & 0.72 (0.0) & 108.3 (0.0) & 2.0 (0.0) & 0.64 (0.0) & 32.8 (0.0) & 6.0 (0.0) & 0.80 (0.0) & 10.5 (0.0) & 7.0 (0.0) \\
        \textbf{\method} & \textbf{0.91 (0.0)} & 1.9 (0.3) & 1.0 (0.2) & \textbf{0.87 (0.0)} & 2.7 (0.1) & 5.0 (0.0) & \textbf{0.88 (0.0)} & 2.7 (0.4) & 6.0 (0.0) \\
    \bottomrule
    \end{tabular}
    \setlength{\tabcolsep}{3pt}
\end{table*}


\section{Additional Analyses}

\subsection{Efficiency of \methodname}

In this part, we analyze the time complexity of our smoothing algorithm, excluding the extracting embeddings from training data and training surrogate model with smoothed data. We first divide our method into two phases and then analyze them individually:
\begin{itemize}[leftmargin=*]
    \item \textbf{Create graph} (\Cref{algo:graph_construction}): Initially, sampling $x$ from $\mathcal V$ and generating Gaussian noise $\epsilon$ are $O(1)$ and $O(d)$ operations, respectively. The interpolation to create $z$ and adding $z$ to $\mathcal V$ are $O(d)$ and $O(1)$. These steps occur in each iteration of the while loop, which runs $N - |\mathcal V|$ times, resulting in a total complexity of $O((N-|\mathcal V|)\cdot d)$. The most significant part of the complexity arises from the edge construction (\Cref{algo:kNN}). Calculating Euclidean distances between $x$ and all other nodes in $\mathcal V$ involves $d$ operations for each of the $|\mathcal V| - 1$ distances, leading to $O(d\cdot |\mathcal V|)$. Retrieving the $K$ closest nodes is $O(|\mathcal V| \log K)$, and constructing the neighborhood around $x$ is $O(K)$. Since these steps must be performed for each node in $\mathcal V$, the overall time complexity for edge construction is $O(|\mathcal V| \cdot (d\cdot |\mathcal V|)) + O(|\mathcal V| \cdot (|\mathcal V|\log K)) + O(|\mathcal V| \cdot K)$, simplifying to $O(d \cdot |\mathcal V|^2)$ since $\log K \ll d$. Given that $|\mathcal V|$ eventually reaches $N$, the overall time complexity is $T_1= O(d\cdot N^2)$.
    \item \textbf{Smooth} (\Cref{algo:smooth}): The algorithm begins with constructing the weighted adjacency matrix $A$ as defined in \Cref{eq:weighted-adj}. This construction involves calculating the Euclidean distance between each pair of nodes. Since this computation has been done in the previous step, the time complexity of this step is $O(N^2)$. As shown in Equation \ref{eq:label_propagation}, a label propagation layer calculates the inverse square root of the degree matrix, $D^{-1/2}$, and multiplies it by the adjacency matrix, $A$. Efficiently, the inverse of the diagonal matrix D can be computed in linear time, $O(N)$. According to \citet{near_linear}, label propagation can be interpreted as label exchange across edges, resulting in a time complexity of $O(|\mathcal E|)$, significantly less than the naive $O(N^3)$ matrix multiplication. Thus, the computation time involves $O(N^2)$ for constructing $A$ and $D$, $O(N)$ for taking the inversion, and $O(m \cdot |\mathcal E|)$ for running label propagation in $m$ layers. Moreover, since kNN graphs exhibit sparse structures with numerous sub-communities, the adjacency matrix $A$ is also sparse, and thus $O(m \cdot |\mathcal E|) \ll O(m \cdot N^2)$ in our work. Therefore, the overall time complexity is $T_2 = 2O(N^2) + O(N) + O(m \cdot |\mathcal E|) \approx O(N^2)$.

    
\end{itemize}
In summary, \method has a time complexity of $T = T_1 + T_2 = O(d \cdot N^2) + O(N^2)$. Since $O(N^2)$ is dominated by $O(d \cdot N^2)$, $T$ simplifies to $O(d \cdot N^2)$. 

\begin{table*}[t]
\centering
\caption{Results of best selected hyperparameters for each task.}{
    \label{tab:app-hparams}
    \begin{tabular}{lc|ccccc|ccc}
        \toprule
        Task & Difficulty & $N$ & $m$ & $\alpha$ & $k$ & Algorithm & Fitness & Diversity & Novelty \\
        \midrule
        \multirow{3}{*}{AAV} & Harder1 & 4000 & 1 & 0.6 & 4 & \multirow{3}{*}{L-BFGS} & 0.56 (0.0) & 6.2 (1.0) & 13.0 (0.0) \\
        & Harder2 & 4000 & 1 & 0.6 & 4 & & 0.51 (0.0) & 6.2 (1.6) & 12.8 (0.5) \\
        & Harder3 & 4000 & 1 & 0.6 & 4 & & 0.45 (0.1) & 9.4 (1.6) & 13.2 (0.5) \\
        \midrule
        \multirow{3}{*}{GFP} & Harder1 & 18000 & 4 & 0.2 & 4 & \multirow{3}{*}{Grad. Ascent} & 0.89 (0.0) & 3.0 (0.2) & 7.4 (0.6) \\
        & Harder2 & 14000 & 4 & 0.2 & 5 & & 0.87 (0.0) & 3.1 (0.2) & 7.6 (0.6) \\
        & Harder3 & 14000 & 4 & 0.2 & 4 & & 0.78 (0.1) & 5.2 (2.4) & 8.0 (0.0) \\
        \bottomrule
    \end{tabular}
}
\end{table*}

\section{Hyperparameters Tuning Process}\label{app:method-hparams}

For all tasks detailed in the main text, we keep the architecture of the VAE and the surrogate model, as well as the hyperparameters for the optimization algorithms, consistent. We then use the Optuna \cite{10.1145/3292500.3330701} package to tune the hyperparameters for each task. The range of hyperparameters is listed below:
\begin{itemize}
    \item Number of nodes $N: \{ 
4000,5000,6000,\ldots,20000 \}$;
    \item Coefficient $\alpha: \{ 0.1, 0.15, 0.2, \ldots, 0.9 \}$;
    \item Number of propagation layers $m:\{1, 2, 3, 4\}$;
    \item Number of neighbors $k:\{ 2, 3, 4, \ldots, 8 \}$
\end{itemize}

\Cref{tab:app-hparams} presents the results of our method with the best corresponding hyperparameters for each main task. It is clear that the AAV dataset requires a smaller graph size, with 4,000 nodes and fewer propagation layers ($m=1$) to optimize effectively. Conversely, the GFP dataset achieves better performance with a larger graph size, with nodes ranging from 14,000 to 18,000 and four propagation layers. The other hyperparameters, $\alpha$ and $k$, remain consistent across different difficulties within the same dataset. These findings indicate that while hyperparameter selection is highly dependent on the landscape characteristics of each dataset, it remains stable across varying difficulty levels within the same dataset, thereby reducing the effort needed to find optimal settings.

\section{Design-Bench}\label{app:design-bench}

\paragraph{\textbf{Tasks}} To demonstrate the domain-agnostic nature of our method, we conduct experiments on three tasks from Design-Bench\footnote{We exclude domains with highly inaccurate, noisy oracle functions (ChEMBL, Hopper, Superconductor, and TF Bind 10) or those too expensive to evaluate (NAS).} \cite{pmlr-v139-trabucco21a}. \textbf{D'Kitty} and \textbf{Ant} are continuous tasks with input dimensions of 56 and 60, respectively. The goal is to optimize the morphological structure of two simulated robots: Ant \cite{brockman2016openaigym} to run as fast as possible, and D'Kitty \cite{pmlr-v100-ahn20a} to reach a fixed target location. \textbf{TF Bind 8} is a discrete task, where the objective is to find the length-8 DNA sequence with maximum binding affinity to the \texttt{SIX6\_REF\_R1} transcription factor. The design space consists of sequences of one of four categorical variables, corresponding to four types of nucleotides. For each task, Design-Bench provides a \texttt{public} dataset, a larger \texttt{hidden} dataset for score normalization, and an exact oracle to evaluate proposed designs. To simulate the labeled scarcity scenarios, we randomly subsample $1\%$ of data points in the \texttt{public} set of each task.

\paragraph{\textbf{Baselines}} We compare \method with four state-of-the-art models in the offline setting: MINs \cite{NEURIPS2020_373e4c5d}, BONET \cite{pmlr-v202-mashkaria23a}, BDI \cite{NEURIPS2022_bd391cf5}, and ExPT \cite{NEURIPS2023_8fab4407}. All baseline results are sourced from those reported by \citet{NEURIPS2023_8fab4407}.

\paragraph{\textbf{Evaluation}} For each method, we allow an optimization budget of $Q=256$. We report the median, max, and mean scores among the 256 proposed inputs. Following prior works, scores are normalized to $[0, 1]$ using the minimum and maximum function values from a large hidden dataset: $y_{\text{norm}} = \frac{y - y_{\min}}{y_{\max} - y_{\min}}$. The mean and standard deviation of the scores are reported across three independent runs for each method.

\paragraph{\textbf{Implementation Details}}
For each task, we use the full \texttt{public} dataset to train the VAE, which consists of a 6-layer Transformer encoder with 8 attention heads and a latent dimension size of 128. The decoder is a deep convolutional neural network, as described in \Cref{sec:vae}. For model-based optimization, we utilize Gradient Ascent with learning rate of $0.005$ for $500$ iterations. For latent graph-based smoothing, we use the same hyperparameter ranges listed in \Cref{app:method-hparams} and employ Optuna to tune the hyperparameters for each task. \Cref{tab:app-db-hparams} outlines the hyperparameter selection for each task.

\begin{table}[t]
    \centering
    \caption{Optimal settings of Design-Bench tasks.}
    \label{tab:app-db-hparams}
    \begin{tabular}{c|cccc}
        \toprule
        Task & $N$ & $m$ & $\alpha$ & $k$ \\
        \midrule
        D'Kitty & 16000 & 3 & 0.3 & 5 \\
        Ant & 16000 & 3 & 0.3 & 5 \\
        TF Bind 8 & 14000 & 6 & 0.6 & 2 \\
        \bottomrule
    \end{tabular}
\end{table}

\end{document}